\documentclass[10pt,twocolumn,letterpaper]{article}

\usepackage{cvpr}              

\usepackage{algorithm}
\usepackage{algpseudocode}
\usepackage{amsmath}
\usepackage{booktabs}
\usepackage{dsfont}
\usepackage{graphicx}
\usepackage{makecell}
\usepackage{multirow}
\usepackage{pifont}
\usepackage[group-separator={,}]{siunitx}
\usepackage{tabularx}
\usepackage{xcolor}
\usepackage{caption}
\usepackage{soul}
\usepackage{changepage,threeparttable}
\usepackage{colortbl}
\usepackage{mathtools}
\usepackage{xfrac}
\usepackage{catchfile}
\usepackage{longtable}
\usepackage{tabu}
\usepackage{supertabular}
\usepackage{multicol}
\usepackage[accsupp]{axessibility}  

\newcolumntype{Y}{>{\centering\arraybackslash}X}
\newcommand{\quotes}[1]{``#1''} 

\definecolor{darkred}{rgb}{0.7,0.2,0.1}
\definecolor{darkgreen}{rgb}{0,0.7,0}
\definecolor{orange}{RGB}{255,127,0}
\definecolor{ourpurple}{RGB}{127,127,204}
\definecolor{palgreen}{RGB}{51,179,179}
\definecolor{magenta}{RGB}{199,21,133}
\definecolor{color_1}{RGB}{255,0,128}
\definecolor{color_2}{RGB}{128,128,0}
\definecolor{color_3}{RGB}{0,128,0}
\definecolor{color_4}{RGB}{128,0,0}
\definecolor{color_5}{RGB}{128,0,128}
\definecolor{cadetgrey}{RGB}{0.57, 0.64, 0.69}
\definecolor{color_red}{RGB}{255,0,0}
\definecolor{color_green}{RGB}{0,255,0}
\definecolor{color_blue}{RGB}{0,0,255}
\definecolor{color_gray}{RGB}{127,127,127}

\newcommand{\projname}[0]{\textup{\textsc{EditCrafter}}}

\newcommand*{\supp}[0]{the \textbf{supplementary material}}

\definecolor{cvprblue}{rgb}{0.21,0.49,0.74}
\usepackage[pagebackref,breaklinks,colorlinks,allcolors=cvprblue]{hyperref}

\def\paperID{70675X} 
\def\confName{CVPR}
\def\confYear{2026}

\title{EditCrafter: Tuning-free High-Resolution Image Editing via Pretrained Diffusion Model}

\author{
Kunho Kim\textsuperscript{1} \quad
Sumin Seo\textsuperscript{2} \quad
Yongjun Cho\textsuperscript{3} \quad
Hyungjin Chung\textsuperscript{4} \\[0.2em]
\textsuperscript{1}NC AI $\quad$ \textsuperscript{2}Medipixel, Inc. $\quad$
\textsuperscript{3}MAUM.AI $\quad$
\textsuperscript{4}EverEx
\\[0.4em]
\textcolor{magenta}{\url{https://editcrafter.github.io}}
}

\begin{document}

\twocolumn[{%
\renewcommand\twocolumn[1][]{#1}%

\maketitle

\vspace{-27.5pt}
\begin{center}
    \captionsetup{type=figure, labelfont=bf}
    \includegraphics[width=\linewidth]{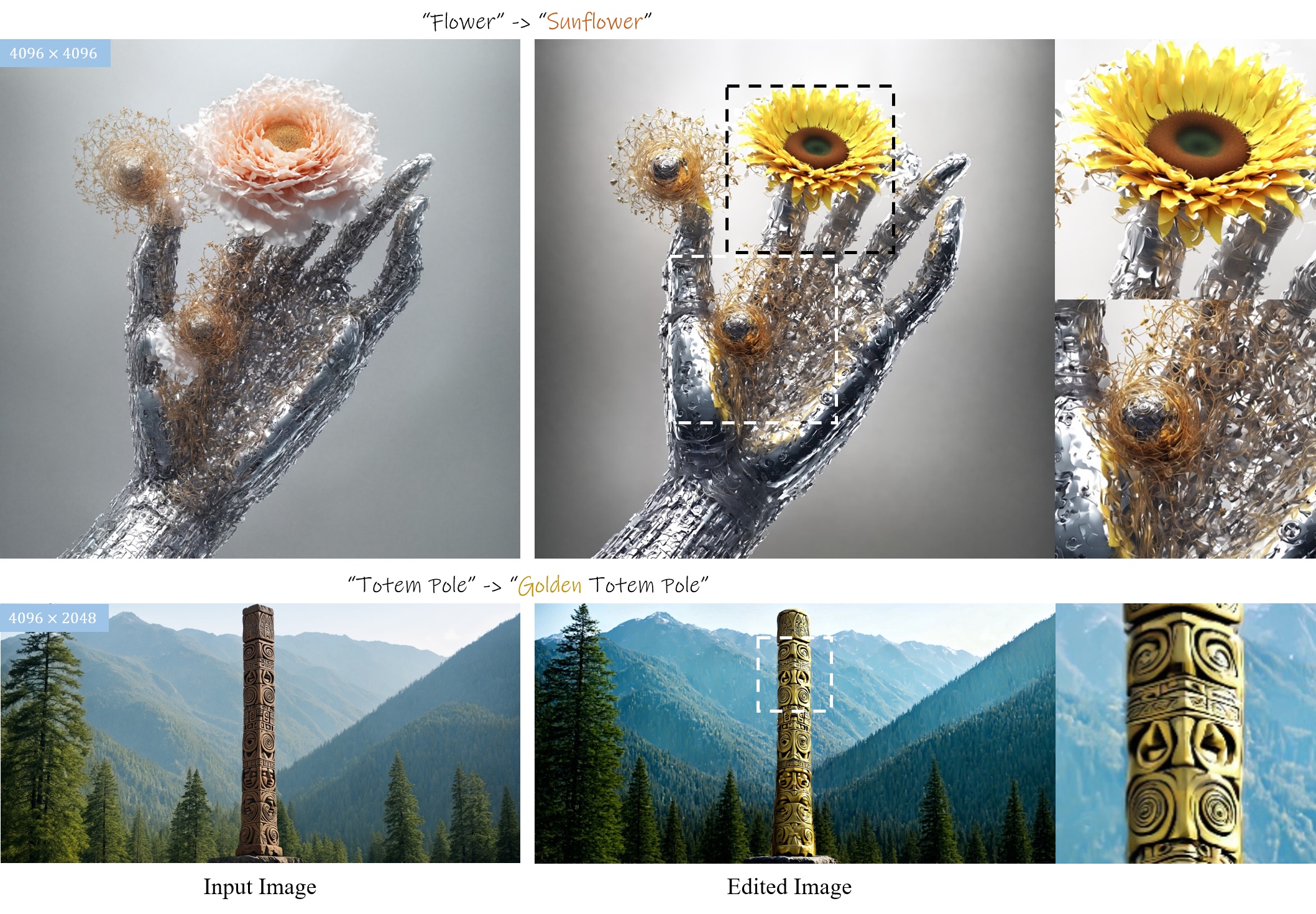}
    \vspace{-20pt}
    \caption{
    Our proposed framework, \projname{}, facilitates text-guided image editing at resolutions up to 4K while meticulously preserving the high-resolution details of the input images using only a single editing prompt.
    }
    \label{fig:teaser}
\end{center}
}]


\begin{abstract}
We propose ~\projname{}, a high-resolution image editing method that operates without tuning, leveraging pretrained text-to-image (T2I) diffusion models to process images at resolutions significantly exceeding those used during training.
Leveraging the generative priors of large-scale T2I diffusion models enables the development of a wide array of novel generation and editing applications. 
Although numerous image editing methods have been proposed based on diffusion models and exhibit high-quality editing results, they are difficult to apply to images with arbitrary aspect ratios or higher resolutions since they only work at the training resolutions (512$\times$512 or 1024$\times$1024). 
Naively applying patch-wise editing fails with unrealistic object structures and repetition. To address these challenges, we introduce ~\projname{}, a simple yet effective editing pipeline. ~\projname{} operates by first performing tiled inversion, which preserves the original identity of the input high-resolution image. We further propose a noise-damped manifold-constrained classifier-free guidance (NDCFG++) that is tailored for high resolution image editing from the inverted latent. Our experiments show that the our ~\projname{} can achieve impressive editing results across various resolutions without fine-tuning and optimization.
\end{abstract}

\section{Introduction}
\label{sec:intro}
Recent advancements in image synthesis, particularly with text-to-image (T2I) diffusion models trained on large-scale data, have garnered substantial attention from both academia and industry. 
In particular, T2I generation models such as Stable Diffusion ~\cite{Rombach:2022LDM}, SDXL~\cite{Podell:2023SDXL}, Imagen~\cite{Saharia:2022Imagen}, DALLE$\cdot$2~\cite{Ramesh:2022DALLE2}, SD3.5~\cite{Esser:2024SD35}, and FLUX~\cite{flux2024} have gained widespread popularity since its accessibility and notable image quality. 
Despite their success, these models are constrained to resolutions of 512 $\times$ 512 or 1024 $\times$ 1024 since they are trained on those resolutions.

Beyond image synthesis, there has been an emergence of intuitive and powerful text-based editing methods~\cite{Hertz:2023P2P, Cao:2023MasaCtrl, Hila:2023AttendExcite, Han:2024ProxEdit, Mokady:2023NTI, Gaurav:2023Pix2PixZero, Huberman-Spiegelglas:2024EditFriendly, Eyring:2024ReNo, Garibi:2024ReNoise, Deutch:2024TurboEdit, Starodubcev:2024iCD, Nguyen:2025swiftedit, Rout2025RFinversion, flux1kontext2025} for semantically modifying images within pretrained T2I diffusion models, thereby enhancing users' control over the generated content. 
High-resolution image editing is particularly crucial across various domains, including digital content creation and industrial design, where maintaining fine-grained details and structural coherence is essential. 
However, similar to image synthesis, text-guided editing predominantly relies on pretrained diffusion models and therefore inherits limitations related to their training resolutions, which are insufficient for real-world high-resolution applications. 
These limitations pose challenges in applying existing editing methods to such high-resolution scenarios.

To overcome the resolution limitations of pretrained T2I diffusion models, one could directly train on high-resolution images. 
However, increasing the model size and training data scale demands substantial computational resources and extended development time. 
Consequently, previous studies~\cite{Bar2023MultiDiffusion, lee2023SyncDiffusion, quattrini2024MAD} have explored the potential for generating arbitrary-sized images and panoramas, with their latent spaces designed to interact through a patch-wise joint diffusion process due to the inherent limitations of the fixed-resolution of T2I diffusion models. Nevertheless, the field of high-resolution image editing remains relatively underexplored.
Similar to patch-wise high-resolution generation method~\cite{Bar2023MultiDiffusion, lee2023SyncDiffusion, quattrini2024MAD}, one approach~\cite{Kim:2023CSD} utilizes a joint diffusion process with kernel using the pretrained diffusion model to achieve consistency across a set of patch images. 
However, this approach often struggles to the object repetition issue as shown in Fig.~\ref{fig:main_comparison}.
This issue of object repetition arises from guiding each patch with the same text prompt using classifier-free guidance~\cite{Ho:2022CFG}, even when the object is absent in the patch. 
One naive approach is to independently apply text-guided image editing to each patch separately and then merge the patches to reconstruct the original high-resolution image. 
However, dividing a high-resolution image into local patches may result in each patch containing only a partial representation of the object rather than the full object, and the content of the patch may not align with the text prompt.

Building upon these observations, we introduce a tuning-free high-resolution image editing strategy, \projname{}. 
We carefully analyze the reconstruction capabilities of the inversion method in its patch-wise implementation to determine how to preserve the original information of the input high-resolution image. 
By employing appropriate tiled inversion techniques, we obtain high-resolution latent representations that are conducive to editing.

A subsequent challenge is guiding these high-resolution latents with a single text prompt to achieve high-quality edits. To address this, we adopt the high-resolution image generation method~\cite{he2023scalecrafter}, which replaces the standard convolution layers with dilated convolution layers throughout the entire U-Net, utilizing the pretrained parameters as the editing framework.
Furthermore, to effectively utilize the generator as an editor, we propose a manifold-constrained noise-damped classifier-free guidance (NDCFG++) approach. This sampling process enhances the ability to guide the high-resolution latents accurately, ensuring high-quality image edits aligned with the provided text prompts.

To investigate the effectiveness of our proposed approach, we conduct extensive experiments on high-resolution image editing, evaluating our method both quantitatively and qualitatively.
Ablation studies demonstrate the superiority of our method over a direct approach including super-resolution upsampler with editing method using down-sized input images.
We curated image-text pairs using a high-resolution generation model~\cite{ren2024ultrapixel} to quantitatively evaluate our method, showing that ours outperforms baseline approaches in terms of human preference and image-text alignment, and further indicating that our design choice yields superior results.
Qualitative results illustrate that our method effectively modifies the target object while avoiding the object repetition, achieving strong alignment with the text prompt.
This improvement in inversion strategy and text guidance adherence highlights the contribution of our approach, which better follows the specified instructions compared to a previous method~\cite{Kim:2023CSD}, without visible seams between patches or unwanted object repetition. 

\section{Related Work}
\label{sec:related_work}

\subsection{Text-to-Image Diffusion Models}
Diffusion models~\cite{Ho:2020DDPM, Rombach:2022LDM} have demonstrated remarkable success in generating high-quality images, providing a foundation for later advancements in various image generation tasks including text-guided image generation~\cite{Nichol:2022Glide, Saharia:2022Imagen,Ramesh:2022DALLE2} and image-to-image translation~\cite{Isola:2018Pix2Pix, Avrahami:2023BLD, Zhang2023ControlNet, Kwon:2022CLIPstyler}.
Recent work has extended T2I diffusion models to various domains, including video generation~\cite{Ho:2023VDM, Ceylan:2023Pix2Video} and 3D generation~\cite{Poole:2023Dreamfusion, Lin:2023Magic3d, Shue:2023TriplaneDiffusion, Koo:2023SALAD}, where the focus lies in ensuring temporal and spatial consistency across generated frames and structures.
Furthermore, ControlNet~\cite{Zhang2023ControlNet} has introduced end-to-end architecture for integrating various conditioning inputs—such as sketches, depth maps, or poses—within the diffusion process, thereby enhancing the training efficiency and adaptability of large-scale T2I models to follow complex editing conditions.
Our method enables higher-resolution image editing which utilizes the generation capability of pretrained large-scale T2I diffusion models~\cite{Rombach:2022LDM}.
\subsection{High-resolution Image Generation}
High-resolution image generation has progressed through efforts to address key challenges, including computational inefficiency, high training costs, and structural artifacts. 
Training models directly on high-resolution images~\cite{chen2024pixartalpha, chen2024pixartsigma, ren2024ultrapixel, xie2025sana} is indeed a viable approach. However, this method inherently requires a substantial increase in both model complexity and the volume of training data.
High-resolution image generation also encompasses diffusion models for panoramic view synthesis, which face significant challenges in achieving seamless patch integration and maintaining structural coherence across expansive fields of view~\cite{Bar2023MultiDiffusion, lee2023SyncDiffusion, quattrini2024MAD}.
Latent merging methods~\cite{Bar2023MultiDiffusion, lee2023SyncDiffusion} have introduced a foundational approach to panoramic image generation by employing an averaging technique to smooth transitions between patches, resulting in coherent extended views. 
However, despite achieving high-fidelity panoramas, they still encounters seam artifacts where patches meet, impacting the overall continuity of the image.
To address lack of preserving semantic coherence across intricate scenes and refining structural detail where needed, merging-and-splitting diffusion method~\cite{quattrini2024MAD} builds on an attention-based mechanism that adapts to complex scene structures.
These approaches are inherently limited by the appearance of seams between patches. To address this, we introduce a high-resolution image generation module that integrates patch-wise inverted latents, avoiding the need to generate each patch separately.

Another approach for high-resolution image generation employs kernel dilation~\cite{he2023scalecrafter, huang2024fouriscale, Qiu2025freescale} in conjunction with fixed-size pretrained diffusion models. 
ScaleCrafter~\cite{he2023scalecrafter} introduces a training-free upscaling approach leveraging kernel dilation, which eliminates the need for model retraining. 
Concurrently, FouriScale~\cite{huang2024fouriscale} has applied a frequency-domain perspective to diffusion models with kernel dilation, successfully reducing repetitive patterns and structural distortions, while maintaining visual integrity across scales.
Building on collective advancements in high-resolution image generation~\cite{he2023scalecrafter}, we design a high-resolution editing module adapted from these approaches.








\subsection{Image Editing with Diffusion Models}
Recent studies show that controlling the attention mechanism~\cite{Hertz:2023P2P, Cao:2023MasaCtrl, Hila:2023AttendExcite, Han:2024ProxEdit} in the pretrained diffusion models provide fine-grained control over attention maps to guide the generation process to achieve desired edited details while maintaining the overall consistency of the image.
In parallel, utilizing the inversion process in editing~\cite{Song:2021DDIM, Mokady:2023NTI, Gaurav:2023Pix2PixZero, Huberman-Spiegelglas:2024EditFriendly, Han:2024ProxEdit, Nguyen:2025swiftedit}, which maps a source image to the latent space of the model, enables both generated images from the generation models and real images editing with conditional inputs such as text prompts and reference images. 
Yet, the inversion process is excessively lengthy and underperforms when the data distribution differs from the training data, highlighting the need for investigating text-guidance-free inversion method~\cite{Mokady:2023NTI}. 
However, inversion-based image editing methods are limited by the capabilities of pretrained diffusion models~\cite{Podell:2023SDXL, Rombach:2022LDM}, which significantly restrict high-resolution image editing.

To the best of our knowledge, CSD~\cite{Kim:2023CSD} is the first to achieve high-resolution image editing, utilizing a score distillation approach for synchronous patch-wise generation. 
CSD proposes score functions based on the Stein Variational Gradient Descent (SVGD) method, which are applied across patches to enable more coherent generative priors for high-resolution images, including panoramic views.
However, boundary artifacts between patches remain a challenge due to the quality of patch-wise sampling and are limited by the dependency on pretrained models~\cite{Brooks:2023InstructPix2Pix}. 
Our study addresses this limitation by incorporating patch-wise editing along with high-resolution generation models, allowing seamless, artifact-free editing across image patches, suggested with extensive high-quality image editing results on qualitative analysis.

\begin{figure*}[!t]
\captionsetup{type=figure, labelfont=bf}
\centering
\includegraphics[width=0.9\textwidth]{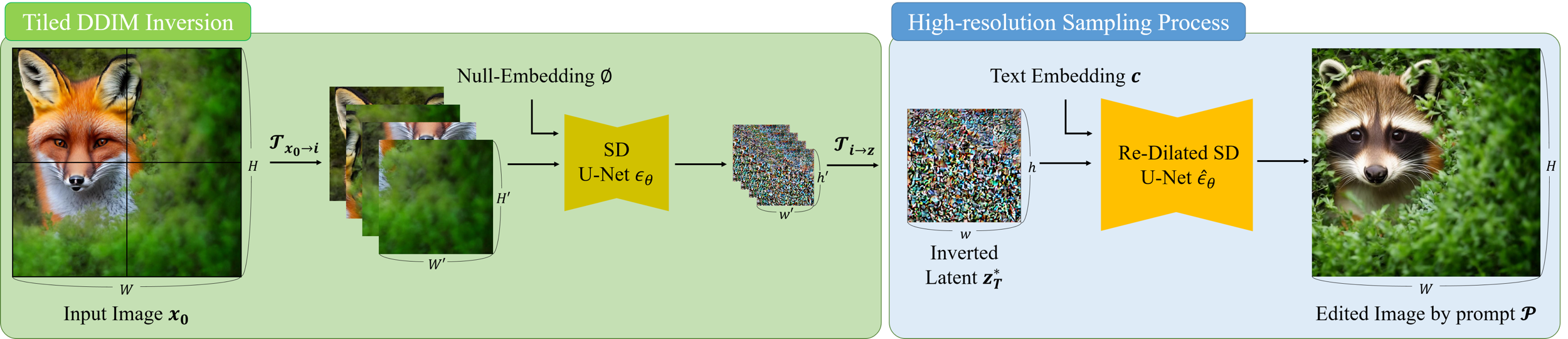}
\vspace{-5pt}
\caption{
The overview of \projname{} pipeline. Since direct inversion of high-resolution images using the pretrained Stable Diffusion (SD) model is not feasible, we first perform tiled DDIM inversion to generate a high-resolution latent representation. Utilizing this latent, the reverse diffusion process is carried out with a re-dilated noise estimator. To enhance the quality of text-guided editing, we propose manifold-constrained noise-damped classifier-free guidance (NDCFG++). In this figure, editing prompt $\mathcal{P}$ is \textit{\quotes{A raccoon peeking out from behind a bush}}.
}
\label{fig:pipeline}
\end{figure*}


\section{Method}
\label{sec:method}
Our goal to edit the high-resolution image $x_0 \in \mathbb{R}^{H \times W \times 3}$ using the text prompt $\mathcal{P}$ with a text-guided diffusion model~\cite{Rombach:2022LDM, Podell:2023SDXL} which are trained by the fixed-size low-resolution image $x_0' \in \mathbb{R}^{H' \times W' \times 3}$. 
To perform such editing operations, it is essential to first invert the image $x_0$ into the latent domain. 
The primary challenge lies in accurately inverting a high-resolution image $x_0$ with the fixed-size pretrained T2I diffusion model, while simultaneously preserving the model's intuitive text-based editing capabilities.

Diffusion models are typically trained on resolutions of $512 \times 512$ or $1024 \times 1024$, which limits their capability to directly invert high-resolution images $x_0$. 
To address this limitation, we propose a straightforward yet effective method called tiled DDIM inversion. 
This approach generates high-resolution latent that is amenable to editing while preserving the original properties of the input high-resolution image $x_0$.

After inverting to obtain an edit-friendly high-resolution latent, an additional challenge emerges: the fixed-size diffusion model cannot handle high-resolution latents. 
To solve this problem, we adopt the high-resolution image generator~\cite{he2023scalecrafter}, which employs re-dilation techniques to adjust the network's receptive field for higher-resolution images.
However, our observations indicate that directly utilizing the high-resolution image generator does not yield adequate editing capabilities. Therefore, we modify the guidance mechanism of the image generator, enabling it to function effectively as an image editing tool. 

In Sec.~\ref{sec:preliminaries}, we outline the foundational concepts of latent diffusion models. Sec.~\ref{sec:tiled_ddim_inversion} introduces our tiled DDIM inversion technique for generating high-resolution latents. Finally, we describe how to guide the high-resolution latent using fixed-size pretrained diffusion models for editing purposes in Sec.~\ref{sec:sampling}.




\subsection{Preliminaries}
\label{sec:preliminaries}
The latent diffusion model (LDM) uses the encoder $\mathcal{E}$ to encode image $x_0' \in \mathbb{R}^{H' \times W' \times 3}$ into a latent representation $z_0 = \mathcal{E}(x_0') \in \mathbb{R}^{h' \times w' \times c}$ and the decoder $\mathcal{D}$ to reconstruct the image from the latent $\tilde{x}'_0 = \mathcal{D}(z) = \mathcal{D}(\mathcal{E}(x_0'))$. Similar to the denoising diffusion probabilistic models (DDPM)~\cite{Ho:2020DDPM}, LDM executes a forward diffusion process across time steps $t$ ranging from 0 to $T$ in the latent space rather than the image space. 
The forward process that add the noise to the original sample $z_0$ to generate the noisy tractable sample $z_t$ as follows:
\begin{align}
\label{eq:foward_process}
q(z_t|z_0)=\mathcal{N}(z_t; \sqrt{\bar{\alpha}} z_0, (1 - \bar{\alpha}_t) I),
\end{align}
where $\alpha_t$ is a predefined variance scheduler and $\bar{\alpha} = \prod_{i=1}^{t} \alpha_i$. After the sufficient time steps $T$, $q(z_t|z_0)$ converges to a unit Gaussian $\mathcal{N}(0,I)$. 
The reverse process gradually remove the noise and predict a clean sample $z_{t-1}$ from the previous sample $z_t$:
\begin{align}
\label{eq:reverse_process}
p_\theta(z_{t-1}|z_t)=\mathcal{N}(z_{t-1}; \mu_\theta(z_t, t), \beta_t I),
\end{align}
where $\mu_\theta$ and $\beta_t$ represent the mean and the time-dependent constant variance, respectively.
The mean $\mu_\theta(z_t, t) = \frac{1}{\sqrt{\alpha_t}}\left(z_t - \frac{\beta_t}{\sqrt{1- \bar{\alpha}_t}}\epsilon\right)$ can be calculated by the noise estimator predicting the noise $\epsilon$ given $z_t$.

The goal is training the noise estimator (U-Net) to predict the noise $\epsilon$ from the noisy latents $z_t$. In text-guided diffusion models, we can control the reverse process through input a text embedding $c$ of a text prompt $P$, which is obtained using a text encoder such as CLIP~\cite{Radford:2021CLIP}. 


Since Stable Diffusion (SD) is trained with text conditioning, we can modulate the influence of the conditional text prompt $\mathcal{P}$ during the reverse diffusion process by employing classifier-free guidance (CFG)~\cite{Ho:2022CFG}.
Utilizing the null-text embedding $\varnothing$ extracted from a null-text \quotes{} and the guidance scale parameter $\omega$, the noise prediction incorporating classifier-free guidance is computed as follows:
\begin{align}
\label{eq:cfg_reverse}
\epsilon_{\theta}(z_t, t, c, \varnothing) &= \epsilon_{\theta}(z_t, t, \varnothing) \notag \\
&+ \omega \cdot (\epsilon_{\theta}(z_t, t, c) - \epsilon_{\theta}(z_t, t, \varnothing)),
\end{align}
where the guidance scale $w \geq 0 \in \mathbb{R}$ regulates the relative influence of the conditional prediction $\epsilon_{\theta}(z_t, t, c)$ compared to the unconditional prediction $\epsilon_{\theta}(z_t, t, \varnothing)$.

For simplicity in notation throughout the subsequent sections, we define the conditional prediction $\epsilon_c(z_t) := \epsilon_\theta(z_t, t, c)$ and the unconditional prediction $\epsilon_\varnothing(z_t) := \epsilon_\theta(z_t, t, \varnothing)$. Consequently, Eq.~\ref{eq:cfg_reverse} can be expressed as follows:
\begin{align}
\label{eq:cfg_reverse_simple}
\epsilon_c^\omega(z_t) = \epsilon_{\varnothing}(z_t) + \omega [ (\epsilon_{c}(z_t) - \epsilon_{\varnothing}(z_t)].
\end{align}

The reverse process in SD is inherently stochastic, adhering to the training framework established by DDPM~\cite{Ho:2020DDPM}. 
To accelerate the sampling speed, the deterministic reverse process of the DDIM sampling method~\cite{Song:2021DDIM} is commonly employed. The reverse DDIM process that generates the latent of previous time step $z_{t-1}$ from the current time step latent variable $z_t$ can be written as:
\begin{align}
\label{eq:ddim_reverse}
z_{t-1} = \sqrt{\alpha_{t-1}} z_c^\omega(z_t) + \sqrt{1 - \alpha_{t-1}} \epsilon_c(z_t),
\end{align}
where $z_c^\omega(z_t) = (z_t - \sqrt{1 - \alpha} \epsilon_c^\omega(z_t)) \mathop{/} \sqrt{\alpha_t}$. This reverse process operates in the direction $z_t \rightarrow z_0$.

As detailed in previous work~\cite{Dhariwal2021ADM, Song:2021DDIM}, the reverse DDIM process described in Eq.~\ref{eq:ddim_reverse} is approximately invertible. By rearranging Eq.~\ref{eq:ddim_reverse}, we can derive the DDIM inversion process as follows:
\begin{align}
\label{eq:ddim_inversion}
z_{t+1} = \sqrt{\alpha_{t+1}} z_c^\omega(z_t) + \sqrt{1 - \alpha_{t+1}} \epsilon_c(z_t),
\end{align}
which operates in the direction $z_0 \rightarrow z_t$. 
Consistent with prior work~\cite{Couairon2022DiffEdit, Mokady:2023NTI, Han:2024ProxEdit, Gaurav:2023Pix2PixZero}, we employ this DDIM inversion for the editing process.

\subsection{Tiled DDIM Inversion}
\label{sec:tiled_ddim_inversion}
Since the noise estimator (U-Net) in Stable Diffusion is trained on low-resolution images, directly inverting an encoded high-resolution image $z_0 = \mathcal{E}(x_0)$ into a high-resolution latent $z_t$ for subsequent editing results in poor identity preservation.
To overcome this limitation, we split the high-resolution image into the low-resolution tiles and invert each tile separately. 
Let the tile size $S \in \mathbb{R}^{H' \times W' \times 3}$ match the training resolution of the noise estimator. 
We define the cropping function $\mathcal{T}_{x \rightarrow i}(S): \mathbb{R}^{H \times W \times 3} \rightarrow \mathbb{R}^{H' \times W' \times 3}$ to extract the $i$-th region $x^{(i)}$ from a high-resolution image $x$. 
Its inverse function, $\mathcal{T}_{i \rightarrow x}(S): \mathbb{R}^{H' \times W' \times 3} \rightarrow \mathbb{R}^{H \times W \times 3}$ reintegrates the tiled image into the $i$-th region of the high-resolution image. 
Utilizing the function $\mathcal{T}_{x \rightarrow i}(S)$, high-resolution image is partitioned into non-overlapping tiles with size $S$ thereby facilitating DDIM inversion using Stable Diffusion.
During the inversion of each tile, we minimize the influence of the text condition by setting the CFG guidance scale to $\omega = 0$ in Eq.~\ref{eq:cfg_reverse_simple}, thereby ensuring that $\epsilon_c^\omega(z_t) = \epsilon_{\varnothing}(z_t)$. Consequently, the Eq.~\ref{eq:ddim_inversion} becomes:
\begin{align}
\label{eq:null_inversion}
z_{t+1} = \sqrt{\alpha_{t+1}} z_c^\omega(z_t) + \sqrt{1 - \alpha_{t+1}} \epsilon_\varnothing(z_t)
\end{align}
After inverting each latent $z_T^{(i)*}$, we concatenate all latents to form the high-resolution inverted latent $z_T^*$ using the function $\mathcal{T}_{i \rightarrow z}(S \mathop{/} 8)$. 

This tiled DDIM inversion approach enables the DDIM inversion regardless of input image size as shown in Fig.~\ref{fig:pipeline}. 
Our key observation is that concatenating each inverted latent and creating the high-resolution inverted latent can provide a good initial point for the subsequent reverse process during editing. 
The pseudocode of tiled DDIM inversion is detailed in Alg.~\ref{alg:tiled_inversion}.
\begin{algorithm}
\caption{Tiled DDIM Inversion}
\label{alg:tiled_inversion}

\textbf{Require:} Real image $x_0 \in \mathbb{R}^{H \times W \times 3}$ , Patch size $S \in \mathbb{R}^{H' \times W' \times 3}$
\begin{algorithmic}[1]
\State{$\{ x^{(0)}, x^{(1)}, \cdots, x^{(n)} \} = \mathcal{T}_{x_0 \rightarrow i}(S)$} \Comment{Tiling}
\For{$i=0$ \textbf{to} $n$}
\State{$z_0^{(i)} = \mathcal{E}(x^{(i)})$} \Comment{Encode image}
    \For{$j=0$ \textbf{to} $T-1$}
        \State{$\epsilon_c^\omega(z_t^{(i)}) = \epsilon_\varnothing(z_t^{(i)})$}     
        \State{${z}_c^\omega(z_t^{(i)}) \gets (z_t^{(i)} - \sqrt{1 - \bar{\alpha}_t} \epsilon_\varnothing(z_t^{(i)})) \mathop{/} \sqrt{\bar{\alpha}_t} $}
        \State{$z_{t+1}^{(i)} = \sqrt{\bar{\alpha}_{t+1}} z_c^\omega(z_t^{(i)}) + \sqrt{1 - \bar{\alpha}_{t+1}} \epsilon_c^\omega(z_t^{(i)}) $}
    \EndFor
    \State{$z_T^{(i)*} = \mathcal{T}_{i \rightarrow z}(S \mathop{/} 8)$} \Comment{Mapping}
\EndFor

\State{\Return $z_T^*$}

\end{algorithmic}
\end{algorithm}
\vspace{-10pt}

\subsection{High-resolution Sampling Process with Kernel Dilation}
\label{sec:sampling}
Similar challenges arise during the inversion process, as fixed-size diffusion models are incapable of handling high-resolution latents in the sampling (reverse) process. An existing method~\cite{Kim:2023CSD} addresses this limitation by employing a patch-wise joint reverse process, where overlapping regions between patches interact. 
However, we observe that this approach leads to an unintended issue with object repetition, as it typically relies on a single editing prompt $\mathcal{P}$. This causes SD to propagate the object information specified by the text condition across all patches.
To mitigate this problem, we adopt kernel re-dilation method from ScaleCrafter~\cite{he2023scalecrafter}, which enables seamless high-resolution image generation guided by text. 
A kernel re-dilation technique adjust the network's receptive field for higher-resolution images, ensuring consistency with the receptive field used in the original lower-resolution generation. 
With the dillated kernel, we start the reverse process from a high-resolution latent $z_T \sim \mathcal{N}(0, I_d) \in \mathbb{R}^{h \times w \times c}$ to generate a high-resolution image $x_0 \in \mathbb{R}^{H \times W \times 3}$. 

Increasing the convolutional receptive field within dilated blocks impairs the model's denoising capabilities. 
To accurately reconstruct fine structural details while preserving the original denoising performance, noise-damped classifier-free guidance (NDCFG)~\cite{he2023scalecrafter} incorporates a vanilla noise estimator with strong denoising capabilities $\epsilon_\theta$, and a kernel re-dilated noise estimator that generates fine content structures $\tilde{\epsilon}_\theta$ as follows:
\begin{align}
\label{eq:ndcfg}
&\tilde{\epsilon}_c^{\omega}(z_t) = \epsilon_\varnothing(z_t) + \omega [\tilde{\epsilon}_c(z_t) - \tilde{\epsilon}_\varnothing(z_t)] \\
&\tilde{z}_c^{\omega}(z_t) \gets (z_t - \sqrt{1 - \alpha}_t \tilde{\epsilon}_c^\omega(z_t)) \mathop{/} \sqrt{\alpha}_t \\
\label{eq:ndcfg_renoising}
&z_{t-1} = \sqrt{\alpha_{t-1}} \tilde{z}_c^{\omega}(z_t) + \sqrt{1 - \alpha_{t-1}} \tilde{\epsilon}_c^{\omega}(z_t).
\end{align}
When we try to edit with CFG $\omega = 7.5$ as typically used in SD, we observe that it cannot preserve the original input image's information since it was originally devised for the generation purpose. 
Yet, Chung et al.~\cite{Chung:2025CFGpp} has demonstrated that an excessive guidance scale in CFG can degrade the generation quality of T2I diffusion models, whereas their method achieves improved fidelity by reformulating text guidance. 

To integrate the idea that controls detailed information at smaller text guidance scale while preserving object-level information during sampling, we propose manifold-constrained noise-damped classifier-free guidance (NDCFG++) as follows:
\begin{align}
\label{eq:ndcfgpp}
&\tilde{\epsilon}_c^{\lambda}(z_t) = \epsilon_\varnothing(z_t) + \lambda [\tilde{\epsilon}_c(z_t) - \tilde{\epsilon}_\varnothing(z_t)]\\
&\tilde{z}_c^{\lambda}(z_t) \gets (z_t - \sqrt{1 - \alpha_t} \tilde{\epsilon}_c^\lambda(z_t)) \mathop{/} \sqrt{\alpha}_t \\
\label{eq:ndcfgpp_renoising}
&z_{t-1} = \sqrt{\alpha_{t-1}} \tilde{z}_c^{\lambda}(z_t) + \sqrt{1 - \alpha_{t-1}} \epsilon_\varnothing(z_t),
\end{align}
where $\lambda \in [0, 1]$ is a small guidance scale. This NDCFG++ is applied during the initial time steps when $t \leq \tau$, starting from our inverted latent $z_T^{*}$. When $t \geq \tau$, we adhere to the standard CFG++ reverse steps to preserve consistency.

\begin{figure}[!ht]
    \captionsetup{type=figure, labelfont=bf}
    \centering
    \includegraphics[width=0.47\textwidth]{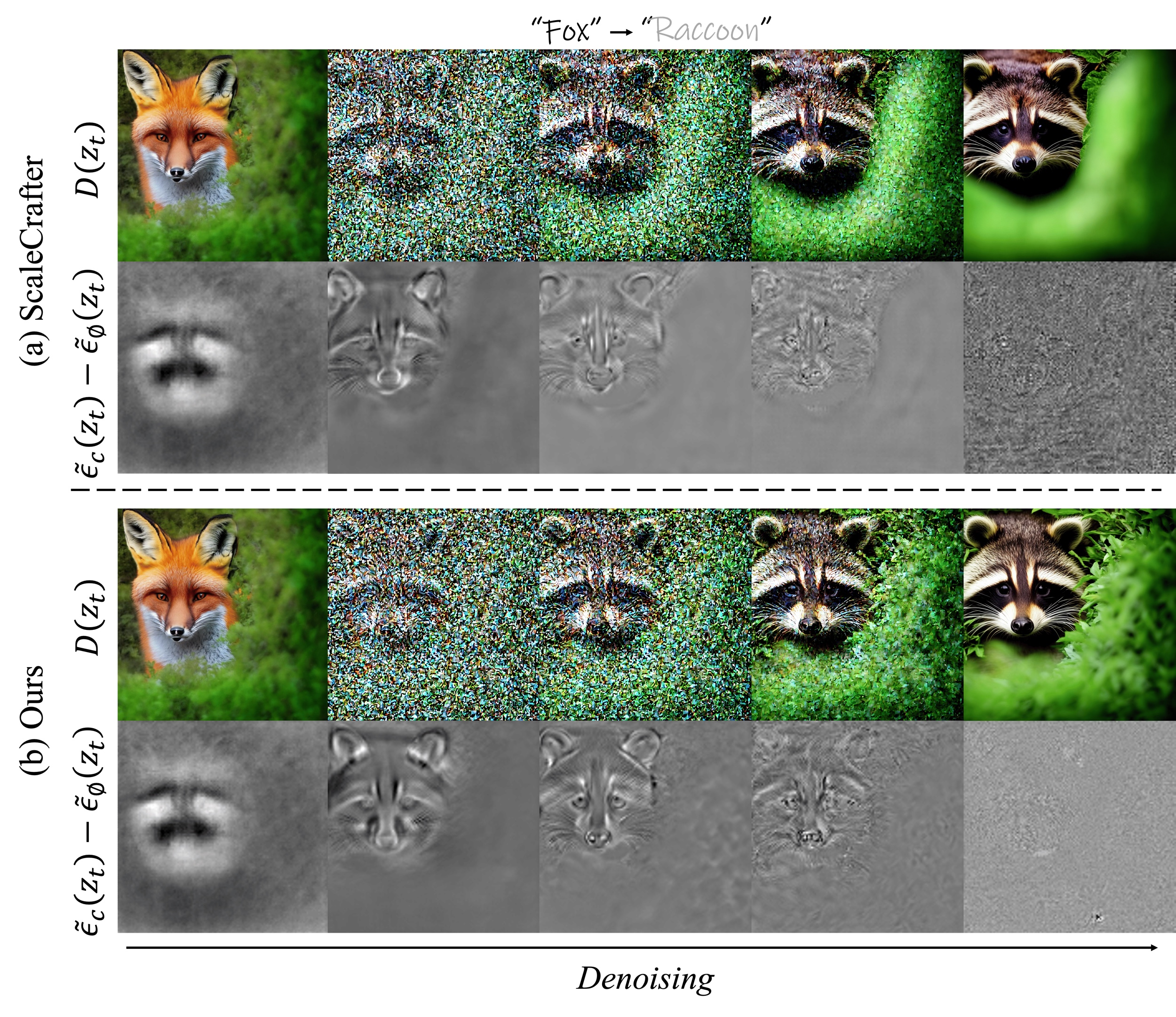}
    \vspace{-10pt}
    \caption{The first and third rows visualize the decoded latents over successive denoising steps. The second and fourth rows show the guidance residual—i.e., the difference between the dilated conditional and unconditional predictions $\epsilon_{c}(z_t)-\epsilon_{\varnothing}(z_t)$. As denoising progresses, our method (NDCFG++) preserves more semantically faithful signal and suppresses background noise, compared with directly applying ScaleCrafter (NDCFG) after tiled inversion.}
    \label{fig:noise_vis}
\end{figure}

Since NDCFG extrapolates beyond and unconditional noise prediction provided by vanilla noise estimator $\epsilon_\theta$ and the difference between conditional and unconditional noise predictions from dilated noise estimator $\tilde{\epsilon}_\theta$ with the large guidance scale $\omega \geq 1$ (Eq.~\ref{eq:ndcfg}), the resulting estimates may potentially deviate from the data manifold.
While NDCFG++ interpolates between that unconditional prediction and the difference conditional and unconditional predictions with the small guidance scale $\lambda \in [0, 1]$, the resulting estimates are less likely deviate from the data manifold.
Additionally, a key distinction between NDCFG and ours NDCFG++ lies in the renoising process (Eq.~\ref{eq:ndcfg_renoising} \& ~\ref{eq:ndcfgpp_renoising}).
Using the unconditional noise $\epsilon_\varnothing(z_t)$ predicted by vanilla noise estimator $\epsilon_\theta$  instead of $\tilde{\epsilon}_c^\omega(z_t)$ provides smoother trajectory of editing as shown in Fig.~\ref{fig:noise_vis}.
The whole algorithm of our sampling process is outlined in Alg.~\ref{alg:reverse_diffusion_ours}.



    


\begin{algorithm}[!hb]
\caption{Reverse Diffusion with Ours}
\label{alg:reverse_diffusion_ours}

\textbf{Require:} Inverted latent $z_T^*$,  $\lambda \in [0, 1]$, $\tau \leq T \in \mathbb{R}$  
\begin{algorithmic}[1]
\For{$t= T$ \textbf{to} $1$} 
    \If{$t \leq \tau$} \Comment{NDCFG++}
        \State{$\tilde{\epsilon}_c^{\lambda}(z_t^{*}) = \epsilon_\varnothing(z_t^{*}) + \lambda [\tilde{\epsilon}_c(z_t^{*}) - \tilde{\epsilon}_\varnothing(z_t^{*})]$} 
        \State{$\tilde{z}_c^{\lambda}(z_t^{*}) \gets (z_t^{*} - \sqrt{1 - \alpha_t} \tilde{\epsilon}_c^\lambda(z_t^{*})) \mathop{/} \sqrt{\alpha}_t $}
        \State{$z_{t-1}^{*} = \sqrt{\alpha_{t-1}} \tilde{z}_c^{\lambda}(z_t^{*}) + \sqrt{1 - \alpha_{t-1}} \epsilon_\varnothing(z_t^{*}) $}

    \Else \Comment{Vanilla CFG++}
        \State{$\tilde{\epsilon}_c^{\lambda}(z_t^{*}) = \tilde{\epsilon}_\varnothing(z_t^{*}) + \lambda [\tilde{\epsilon}_c(z_t^{*}) - \tilde{\epsilon}_\varnothing(z_t^{*})]$} 
        \State{$\tilde{z}_c^{\lambda}(z_t^{*}) \gets (z_t^{*} - \sqrt{1 - \alpha}_t \tilde{\epsilon}_c^\lambda(z_t^{*})) \mathop{/} \sqrt{\alpha_t} $}
        \State{$z_{t-1}^{*} = \sqrt{\alpha_{t-1}} \tilde{z}_c^{\lambda}(z_t^{*}) + \sqrt{1 - \alpha_{t-1}} \tilde{\epsilon}_\varnothing(z_t^{*}) $}
    \EndIf
    
\EndFor
\State{$x_0 = \mathcal{D}(z_0^{*})$} \Comment{Decode latent}
\State{\Return $x_0$}
\end{algorithmic}
\end{algorithm}


\begin{figure*}[t]
    \captionsetup{type=figure, labelfont=bf}
    \centering
    \makebox[\textwidth][c]{
        \includegraphics[width=1.0\textwidth]{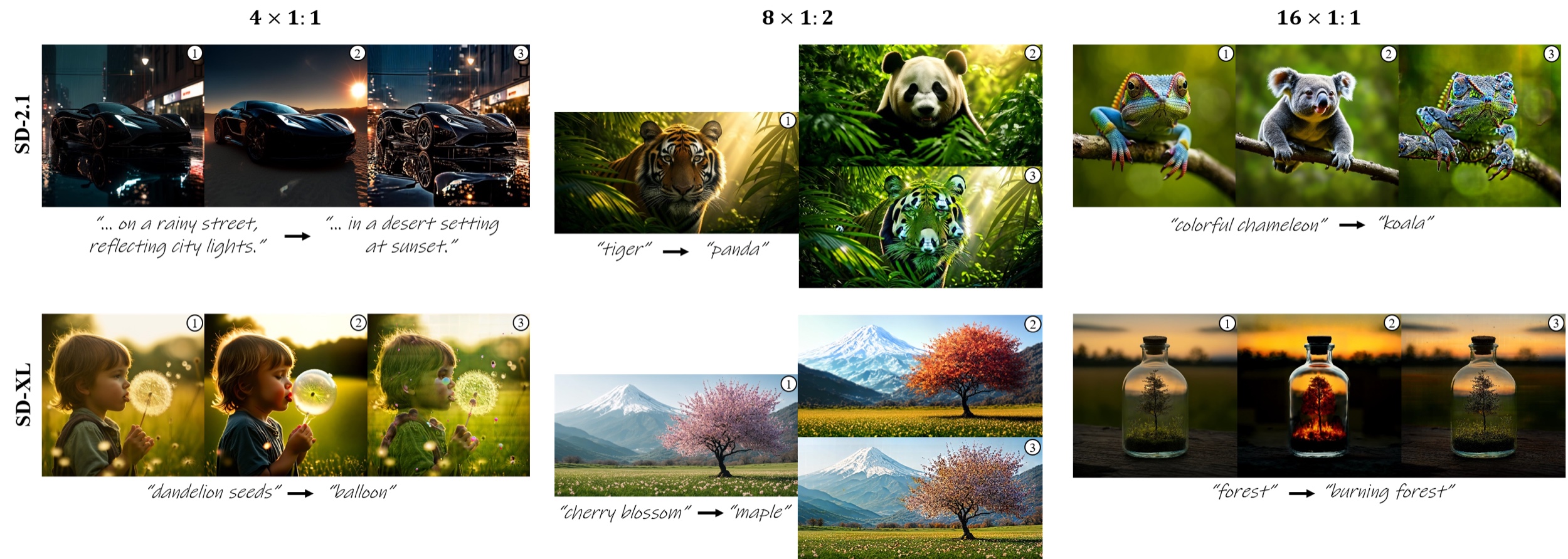}
    }
    \vspace{-20pt}
    \caption{Qualitative comparisons. (1) Original image, (2) Ours, and (3) CSD in 4×, 8× and 16× settings. Best viewed on screen with zoom. The high-quality versions are provided in \supp{}.}
    \label{fig:main_comparison}
\end{figure*}

\section{Experiments}
\label{sec:experiments}

\subsection{Experiment Setup}
\label{sec:exp_setup}

We conduct experiments using two recent pretrained versions of Stable Diffusion, specifically SD 2.1~\cite{Rombach:2022LDM} and SDXL 1.0~\cite{Podell:2023SDXL}, to perform editing at resolutions exceeding their respective training resolutions 512$\times$512, and 1024 $\times$ 1024. 
We scale the number of pixels by factors of 4, 8, and 16, resulting in editing resolutions of 1024$\times$1024, 2048$\times$1024, and 2048$\times$2048 for SD 2.1, and 2048$\times$2048, 4096$\times$2048, and 4096$\times$4096 for SDXL.
All experiments are conducted on a single RTX 4090, demonstrating that our method does not require extensive VRAM (ranging from 3.8GB at 1024×1024 to 18.2GB at 4096×4096).
Unless otherwise specified, all of our experiments are performed using a small guidance scale parameter $\lambda=0.5$. The ablation study on the guidance scale is provided in \supp{}.

\subsubsection{Benchmark}
To evaluate the high-resolution editing performance, we collect an image editing dataset from high-resolution generation model~\cite{ren2024ultrapixel}, known for its high visual fidelity and close resemblance to the natural images. 
The dataset includes 30 images with manually selected text prompts for each resolution including high-resolution square images and wide panoramic images, yielding a total of 150 prompt-image pairs. 
For creating editing prompts, we applied a word-swapping technique to the original prompts used for image generation, replacing nouns that describe the main object or phrases depicting the background.

\subsubsection{Baselines}
To the best of our knowledge, the only existing baseline specifically designed for high-resolution image editing is CSD~\cite{Kim:2023CSD}. 
CSD leverages a joint diffusion approach to generate fixed-size patches using InstructPix2Pix~\cite{Brooks:2023InstructPix2Pix}, a fine-tuned variant of Stable Diffusion.
Throughout the experiments, we use the default parameter settings for CSD.

\subsubsection{Evaluation Metrics}
We utilize HPSv2~\cite{wu2023hpsv2} and ImageReward~\cite{xu2023imagereward} to assess text-to-image alignment based on human preferences. 
Both models are trained on datasets consisting of human preference selections for images corresponding to given text prompts, using 645k and 137k text-image pairs, respectively.
We also evaluate baseline methods using CLIPScore~\cite{hessel2021clipscore} to assess editing quality. CLIPScore measures the similarity between the edited image embedding and the text embedding extracted from the editing prompt $\mathcal{P}$. 

In addition to the metrics mentioned above, user evaluation is a crucial aspect of image editing tasks. Therefore, we conducted a user study using Amazon MTurk to evaluate the effectiveness of our method. A detailed example of the user study is provided in ~\supp{}.

\begin{table}[t]
\captionsetup{type=table, labelfont=bf}
    \centering
    \resizebox{1.0\linewidth}{!}{
    \begin{tabular}{llcccc}
        \toprule
        Model & Res & Method & ImageReward $\uparrow$ & HPSv2 $\uparrow$ & CLIPScore $\uparrow$ \\
        \hline
        \multirow{6}{*}{SD 2.1} 
        & \multirow{2}{*}{4$\times$ 1:1}  & CSD & 0.5538 & 0.2883 & 32.8353 \\
        &  & Ours & \textbf{1.4831} & \textbf{0.2935} & \textbf{34.8039} \\
        \cline{2-6}
        & \multirow{2}{*}{8$\times$ 1:2} & CSD & 0.7165 & 0.2782 & 32.2794 \\
        &  & Ours & \textbf{1.4238} & \textbf{0.2824} & \textbf{34.5303} \\
        \cline{2-6}
        & \multirow{2}{*}{16$\times$ 1:1} & CSD & 0.6304 & 0.2934 & 32.7795 \\
        &  & Ours & \textbf{1.6689} & \textbf{0.3017} & \textbf{35.3194} \\
        \hline
        \multirow{6}{*}{SDXL 1.0} 
        & \multirow{2}{*}{4$\times$ 1:1} & CSD & 0.6304 & 0.2934 & 32.7795 \\
        &  & Ours & \textbf{1.6242} & \textbf{0.2991} & \textbf{34.8067} \\
        \cline{2-6}
        & \multirow{2}{*}{8$\times$ 1:2} & CSD & 0.2939 & 0.2767 & 32.0854 \\
        &  & Ours & \textbf{1.4133} & \textbf{0.2842} & \textbf{33.9795} \\
        \cline{2-6}
        & \multirow{2}{*}{16$\times$ 1:1} & CSD & 0.3699 & 0.2877 & 32.8440 \\
        &  & Ours & \textbf{1.4919} & \textbf{0.2949} & \textbf{34.4959} \\
        \bottomrule
    \end{tabular}
    }
    \vspace{-10pt}
    \caption{Quantitative comparisons.
    }
    \label{tab:quantitative_results}
\end{table}

\subsection{Text-Guided High-Resolution Image Editing}
\label{sec:editing}

\subsubsection{Qualitative Evaluation}
We present qualitative results to compare our method with the baseline in Fig.~\ref{fig:main_comparison}. 
CSD~\cite{Kim:2023CSD} frequently exhibits repetitive objects due to its patch-wise generation scheme. This limitation is illustrated by instances such as pandas appearing on the head of a tiger in edits at 
2048$\times$1024 resolution and koalas spanning the entire body of a chameleon in edits at 2048$\times$2048 resolution.
Furthermore, as the editing resolution increases, CSD tends to adhere to the original image rather than the editing prompt and produce white grids indicating patch boundaries. 
In contrast, our \projname{} demonstrates the capability to modify the shape of the target object or adjust background layout as shown in the edited results for object-centric and panoramic view images, respectively. It consistently generates highly faithful editing images that are well-aligned with the editing prompts while preserving the intricate details of the original images. 
We present more visual editing results in \supp{}.

We also compare our method to a naive baseline for high-resolution image editing, which involves downsampling the image to apply current state-of-the-art editing methods, followed by upsampling to the desired higher resolution using super-resolution techniques. 
To compare ours with state-of-the-art fixed-size editing method, we applied $16\times$ upsampling using StableSR~\cite{wang2024stablesr} on images edited by InfEdit~\cite{Xu:2024InfEdit} which takes the input images resized by $512\times512$ from $2048\times2048$. 
\begin{figure}[t]
    \captionsetup{type=figure, labelfont=bf}
    \centering
    \includegraphics[width=0.48\textwidth]{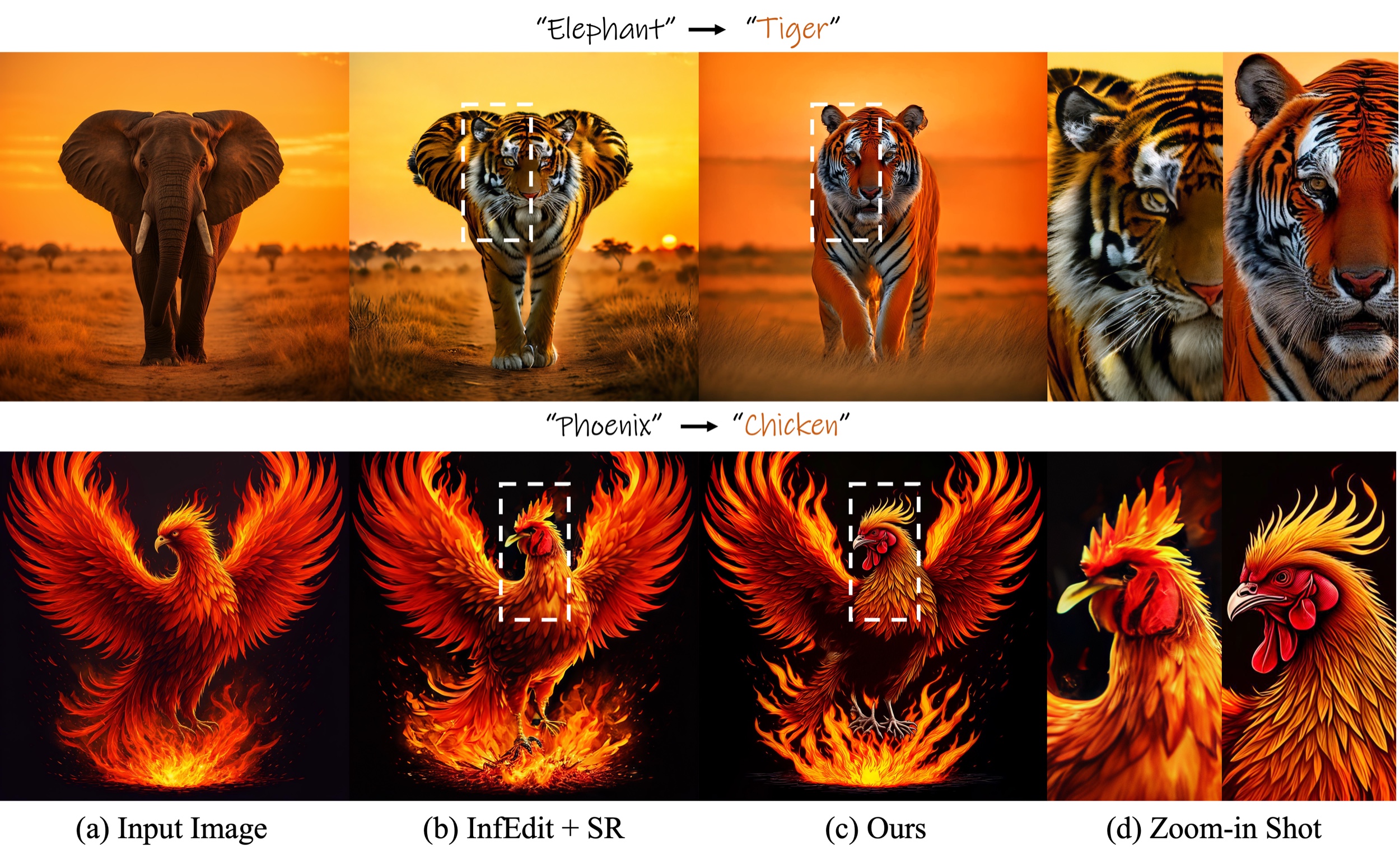}
    \vspace{-20pt}
    \caption{Comparison of $16\times$ super-resolution upsampler~\cite{wang2024stablesr} with InfEdit~\cite{Xu:2024InfEdit} with our approach based on the 16$\times$ SD 2.1. 
    }
    \label{fig:infedit}
\end{figure}


Fig.~\ref{fig:infedit} showcases a qualitative comparison highlighting the superiority of our method, particularly in accurately editing the semantic target while preserving high-resolution detail. 
In both the “Tiger” and “Chicken” examples, our method produces coherent and realistic outputs with structural integrity and fine-grained texture details, most notably in the eye regions and furs. In contrast, InfEdit + StableSR produces distorted facial features, indicating that performing edits at low resolution followed by upsampling leads to a loss of fine-grained details.
Comprehensive comparisons with this approach are provided in \supp{}.

\subsubsection{Quantitative Results}

Tab.~\ref{tab:quantitative_results} represents a quantitative comparison between CSD~\cite{Kim:2023CSD} and \projname{}. 
As shown, \projname{} demonstrates superior performance over CSD on all resolution. 
The evaluations using ImageReward~\cite{xu2023imagereward}, HPSv2~\cite{wu2023hpsv2}, and CLIPScore~\cite{hessel2021clipscore} demonstrate that our editing results are highly aligned with the editing prompts and human preferences. 
On the other hand, CSD fails to modify the target object as specified by the text prompt, resulting in limited editing responsiveness to object-level changes as shown in the Fig.~\ref{fig:main_comparison}.

For the user study, we collected a total of 25 responses, including 5 vigilance tasks, from 112 participants. The results indicate that human evaluators preferred our \projname{} method in \textbf{72.61\%} of cases compared to CSD. These findings suggest that \projname{} more effectively applies the requested edits, demonstrating stronger alignment with user expectations.

\subsubsection{Ablation Study}
\begin{figure}[t]
    \captionsetup{type=figure, labelfont=bf}
    \centering
    \includegraphics[width=\linewidth]{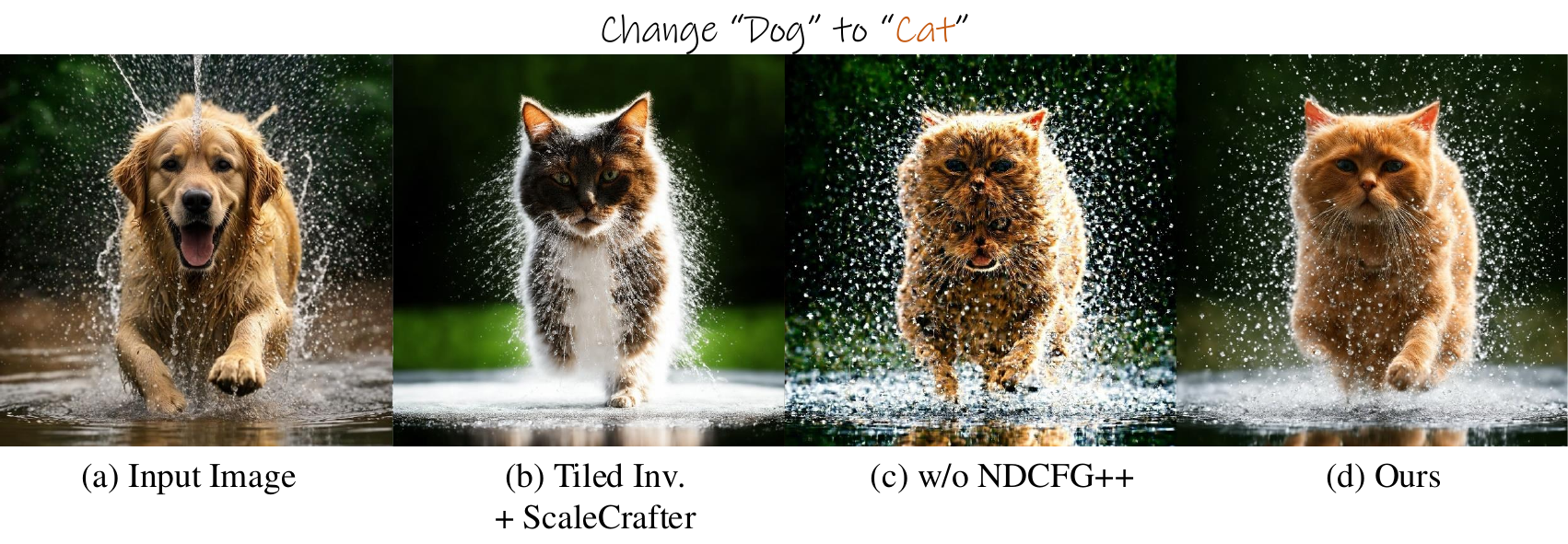}
    \vspace{-20pt}
    \caption{Ablation study qualitative results on the 16$\times$ SD 2.1 . 
    }
    \label{fig:ablation}
    \vspace{-10pt}
\end{figure}

To evaluate contribution of each component in our method, we conducted an ablation study on NDCFG++ within the 16$\times$ SD 2.1 setup. As demonstrated in the second row of Tab.~\ref{tab:ablation_study}, removing NDCFG++ ($\tau = 0$ in Alg.~\ref{alg:reverse_diffusion_ours}) results in performance degradation, with our method achieving the highest performance on text-to-image alignment. Without NDCFG++, performance slightly drops in human preference scores, and CLIP text-image matching score.
As illustrated in Fig.~\ref{fig:ablation}, the contribution of NDCFG++ to qualitative enhancement is evident in its capability in accurately positioning the target object's head in the location of the original object within the source image, ensuring alignment with the object's identity. 

On the one hand, the ablated version of our method, which excludes NDCFG++ and adopts the high CFG generation settings from ScaleCrafter~\cite{he2023scalecrafter}, demonstrates diminished performance as shown in the first row of Tab.~\ref{tab:ablation_study}. 
This decline is evidenced by a substantial reduction in ImageReward scores, as well as decreases in other text-based image quality metrics.
This generation setting overlooks the critical role of text guidance in image editing, adversely affecting both background integrity and the semantic consistency of the object. 
Our method, in contrast, effectively preserves the original background and object textures, including pattern of water droplets or color pattern, resulting in improved fidelity and alignment with the desired text-to-image prompts as shown in the Fig.~\ref{fig:ablation} (d).

\begin{table}[t]
\captionsetup{type=table, labelfont=bf}
    \centering
    \resizebox{1.0\linewidth}{!}{
    \begin{tabular}{ccccc}
    \toprule
    Method & ImageReward $\uparrow$ & HPSv2 $\uparrow$ & CLIP Score $\uparrow$ \\
    \midrule
    Tiled Inv. + ScaleCrafter~\cite{he2023scalecrafter} & 1.2595 & 0.2962 & 34.9431 \\
    Ours w/o NDCFG++ & 1.6273 & 0.2911 & 35.0254 \\
    Ours & \textbf{1.6689} & \textbf{0.3017} & \textbf{35.3194} \\
    \bottomrule
    \end{tabular}
    }
    \caption{Ablation study quantitative results on the 16$\times$ SD 2.1.}
    \label{tab:ablation_study}
\end{table}


\section{Conclusion}
\label{sec:conclusion}
We present \projname{}, a tuning-free and optimization-free editing pipeline using pretrained diffusion models. 
By utilizing an straightforward but effective approach with tiled-inversion and NDCFG++, our method demonstrates consistently superior performance across high-resolution image editing methods.
Notably, we conduct the first extensive quantitative and qualitative evaluation in high-resolution image editing, showcasing the effectiveness of our method's design.
We anticipate that our proposed framework, \projname{}, can be effectively integrated into real-world applications, thereby enhancing the performance in practical settings.

\newpage
\clearpage

\section*{Acknowledgments}

Kunho Kim was supported by Institute for Information \& communications Technology Promotion(IITP) grant funded by the Korea government(MSIT) (RS-2024-00398115, Research on the reliability and coherence of outcomes produced by Generative AI).

\bibliographystyle{ieeenat_fullname}
\bibliography{main}

\newif\ifpaper
\papertrue

\clearpage
\newpage
\appendix

\onecolumn

\setcounter{section}{0}
\renewcommand\thesection{\Alph{section}}
\renewcommand{\thetable}{A\arabic{table}}
\renewcommand{\thefigure}{A\arabic{figure}}
\renewcommand{\theequation}{A\arabic{equation}}
\ifpaper
    \newcommand{\refofpaper}[1]{\unskip}
    \newcommand{\refinpaper}[1]{\unskip}
\else
    \makeatletter
    \newcommand{\manuallabel}[2]{\def\@currentlabel{#2}\label{#1}}
    \makeatother

    \manuallabel{sec:intro}{1}
    \manuallabel{sec:related_work}{2}
    \manuallabel{sec:method}{3}
    \manuallabel{sec:preliminaries}{3.1}
    \manuallabel{sec:tiled_ddim_inversion}{3.2}
    \manuallabel{sec:sampling}{3.3}
    \manuallabel{sec:experiments}{4}

    \manuallabel{eq:foward_process}{1}
    \manuallabel{eq:reverse_process}{2}
    \manuallabel{eq:simple_training_loss}{3}
    \manuallabel{eq:cfg_reverse}{4}
    \manuallabel{eq:cfg_reverse_simple}{5}
    \manuallabel{eq:ddim_reverse}{6}
    \manuallabel{eq:ddim_inversion}{7}
    \manuallabel{eq:null_inversion}{8}
    \manuallabel{eq:ndcfg}{9}
    \manuallabel{eq:ndcfg_renoising}{10}
    \manuallabel{eq:ndcfgpp}{11}
    \manuallabel{eq:ndcfgpp_renoising}{12}

    \manuallabel{fig:teaser}{1}
    \manuallabel{fig:pipeline}{2}
    \manuallabel{fig:main_comparison}{3}
    \manuallabel{fig:infedit}{4}
    \manuallabel{fig:ablation}{5}

    \manuallabel{alg:tiled_inversion}{1}
    \manuallabel{alg:reverse_diffusion_ours}{2}

    \manuallabel{tab:quantitative_results}{1}
    \manuallabel{tab:ablation_study}{2}
    
    \newcommand{\refofpaper}[1]{of the main paper}
    \newcommand{\refinpaper}[1]{in the main paper}
\fi

\section{Implementation Details}
\label{sec:implementation_details}

We provide additional implementation details of Alg.~\ref{alg:reverse_diffusion_ours} \refinpaper{}. 
To highlight the distinguishing factors between ScaleCrafter~\cite{he2023scalecrafter} and our proposed method, we present both reverse processes.
The DDIM sampling steps are configured to $T=50$. 
For $\times$4 editing, we set $\tau=10$ and  for both $\times$8 and $\times$16 editing, $\tau=37$.
These settings are applied to both SD 2.1~\cite{Rombach:2022LDM} and SDXL 1.0~\cite{Podell:2023SDXL}.
We follow the re-dilated convolution configurations for each resolution as implemented in ScaleCrafter.
For both CLIPScore~\cite{hessel2021clipscore} and CLIP Image Similarity~\cite{rinon2022stylegannada}, we employ the \quotes{ViT-B/32} model as the foundational architecture.

\noindent
\begin{minipage}{0.47\textwidth}%
    \begin{algorithm}[H]
    \caption{Reverse Diffusion with ScaleCrafter}
    \label{alg:reverse_diffusion_comp}
    
    \textbf{Require:} $z_T \sim \mathcal{N}(0, \mathbf{I}_d)$,  $0 \leq \omega \in \mathbb{R}$, $\tau \leq T \in \mathbb{R}$  
    \begin{algorithmic}[1]
    \For{$i= T$ \textbf{to} $1$}
        \If{$i \leq \tau$} 
            \State{$\tilde{\epsilon}_c^{\omega}(z_t) = \epsilon_\varnothing(z_t) + \omega [\tilde{\epsilon}_c(z_t) - \tilde{\epsilon}_\varnothing(z_t)]$} 
        \Else
            \State{$\tilde{\epsilon}_c^{\omega}(z_t) = \tilde{\epsilon}_\varnothing(z_t) + \omega [\tilde{\epsilon}_c(z_t) - \tilde{\epsilon}_\varnothing(z_t)]$} 
        \EndIf
        
        \State{$\tilde{z}_c^{\omega}(z_t) \gets (z_t - \sqrt{1 - \alpha}_t \tilde{\epsilon}_c^\omega(z_t)) \mathop{/} \sqrt{\alpha}_t $}
        \State{$z_{t-1} = \sqrt{\alpha_{t-1}} \tilde{z}_c^{\omega}(z_t) + \sqrt{1 - \alpha_{t-1}} \tilde{\epsilon}_c^{\omega}(z_t) $}
    \EndFor
    \State{$x_0 = \mathcal{D}(z_0)$} \Comment{Decode latent}
    \State{\Return $x_0$}
    
    \end{algorithmic}
    \end{algorithm}
\end{minipage}%
\begin{minipage}{0.47\textwidth}%
    \begin{algorithm}[H]
    \caption{Reverse Diffusion with Ours}
    \label{alg:reverse_diffusion_ours_comp}
    
    \textbf{Require:} Inverted latent $z_T^*$,  $\lambda \in [0, 1]$, $\tau \leq T \in \mathbb{R}$  
    \begin{algorithmic}[1]
    \For{$i= T$ \textbf{to} $1$} 
        \If{$i \leq \tau$} \Comment{NDCFG++}
            \State{$\tilde{\epsilon}_c^{\lambda}(z_t^{*}) = \epsilon_\varnothing(z_t^{*}) + \lambda [\tilde{\epsilon}_c(z_t^{*}) - \tilde{\epsilon}_\varnothing(z_t^{*})]$} 
            \State{$\tilde{z}_c^{\lambda}(z_t^{*}) \gets (z_t^{*} - \sqrt{1 - \alpha_t} \tilde{\epsilon}_c^\lambda(z_t^{*})) \mathop{/} \sqrt{\alpha}_t $}
            \State{$z_{t-1}^{*} = \sqrt{\alpha_{t-1}} \tilde{z}_c^{\lambda}(z_t^{*}) + \sqrt{1 - \alpha_{t-1}} \epsilon_\varnothing(z_t^{*}) $}
    
        \Else \Comment{Vanilla CFG++}
            \State{$\tilde{\epsilon}_c^{\lambda}(z_t^{*}) = \tilde{\epsilon}_\varnothing(z_t^{*}) + \lambda [\tilde{\epsilon}_c(z_t^{*}) - \tilde{\epsilon}_\varnothing(z_t^{*})]$} 
            \State{$\tilde{z}_c^{\lambda}(z_t^{*}) \gets (z_t^{*} - \sqrt{1 - \alpha}_t \tilde{\epsilon}_c^\lambda(z_t^{*})) \mathop{/} \sqrt{\alpha_t} $}
            \State{$z_{t-1}^{*} = \sqrt{\alpha_{t-1}} \tilde{z}_c^{\lambda}(z_t^{*}) + \sqrt{1 - \alpha_{t-1}} \tilde{\epsilon}_\varnothing(z_t^{*}) $}
        \EndIf
        
    \EndFor
    \State{$x_0 = \mathcal{D}(z_0^{*})$} \Comment{Decode latent}
    \State{\Return $x_0$}
    
    \end{algorithmic}
    \end{algorithm}
\end{minipage}%

\section{Effect of Classfier-Guidance Scale}
\label{sec:cfg_scale}
\begin{figure*}[!htbp]
    \captionsetup{type=figure, labelfont=bf}
    \centering
    \makebox[\textwidth][c]{
        \includegraphics[width=1.0\textwidth]{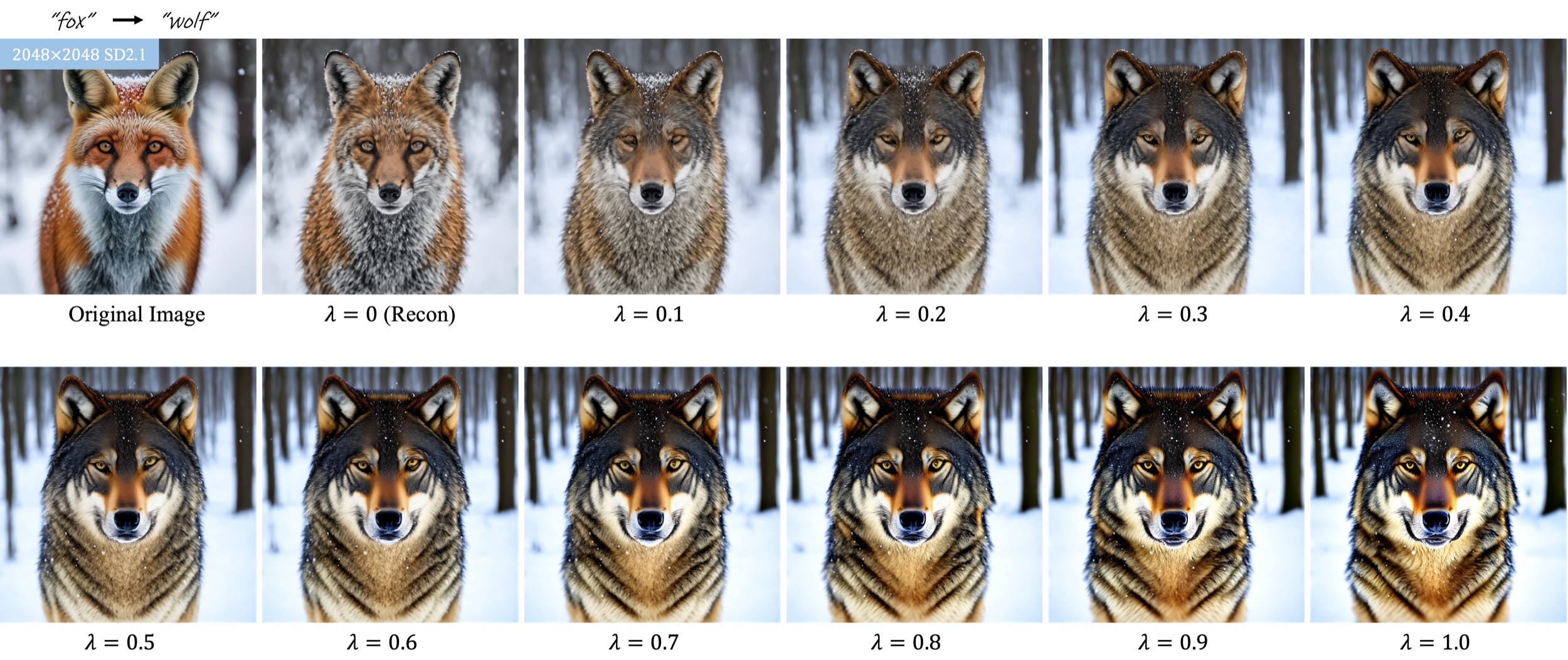}
    }
    \caption{The effect of CFG scale.}
    \label{fig:cfg_scale}
\end{figure*}

\begin{figure*}[!htbp]
    \captionsetup{type=figure, labelfont=bf}
    \centering
    \includegraphics[width=1.0\textwidth]{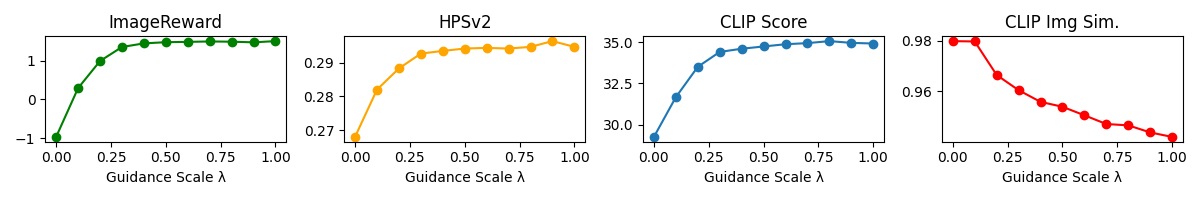}
    \caption{The effect of CFG scale $\lambda$ in 4 $\times$ SD 2.1.}
    \label{fig:cfg_scale_graph}
\end{figure*}

We investigate the effect of small guidance scale $\lambda \in [0, 1]$ in our sampling process. 
We examine the impact of varying the small guidance scale parameter, $\lambda$, within the range [0, 1] on our sampling process. 
As depicted in Fig.~\ref{fig:cfg_scale}, the reconstruction produced with $\lambda=0$ does not exactly replicate the original image; however, it serves as a promising initial foundation for subsequent editing tasks.
Notably, as $\lambda$ increases, the edited images progressively conform more closely to the specified edit prompt \quotes{wolf}. 
This tendency is also reflected when measure the metric. Fig.~\ref{fig:cfg_scale_graph} illustrates that increasing the guidance scale $\lambda$ leads to higher values in edited image-text alignment metrics, while simultaneously reducing the preservation of the original image as measured by CLIP Image Similarity~\cite{rinon2022stylegannada}.

This behavior indicates that higher guidance scales enhance the alignment between the generated modifications, thereby facilitating more precise and controlled image editing.
Based on our observations, we set the guidance scale parameter $\lambda=0.5$ to achieve an optimal balance between adhering to the editing prompt and preserving the original identity for all experiments. 
However, users may adjust this setting to better suit real-world editing applications.


\newpage
\section{User Study}
\label{sec:user_study}
\begin{figure*}[!ht]
    \captionsetup{type=figure, labelfont=bf}
    \centering
    \includegraphics[width=0.8\textwidth]{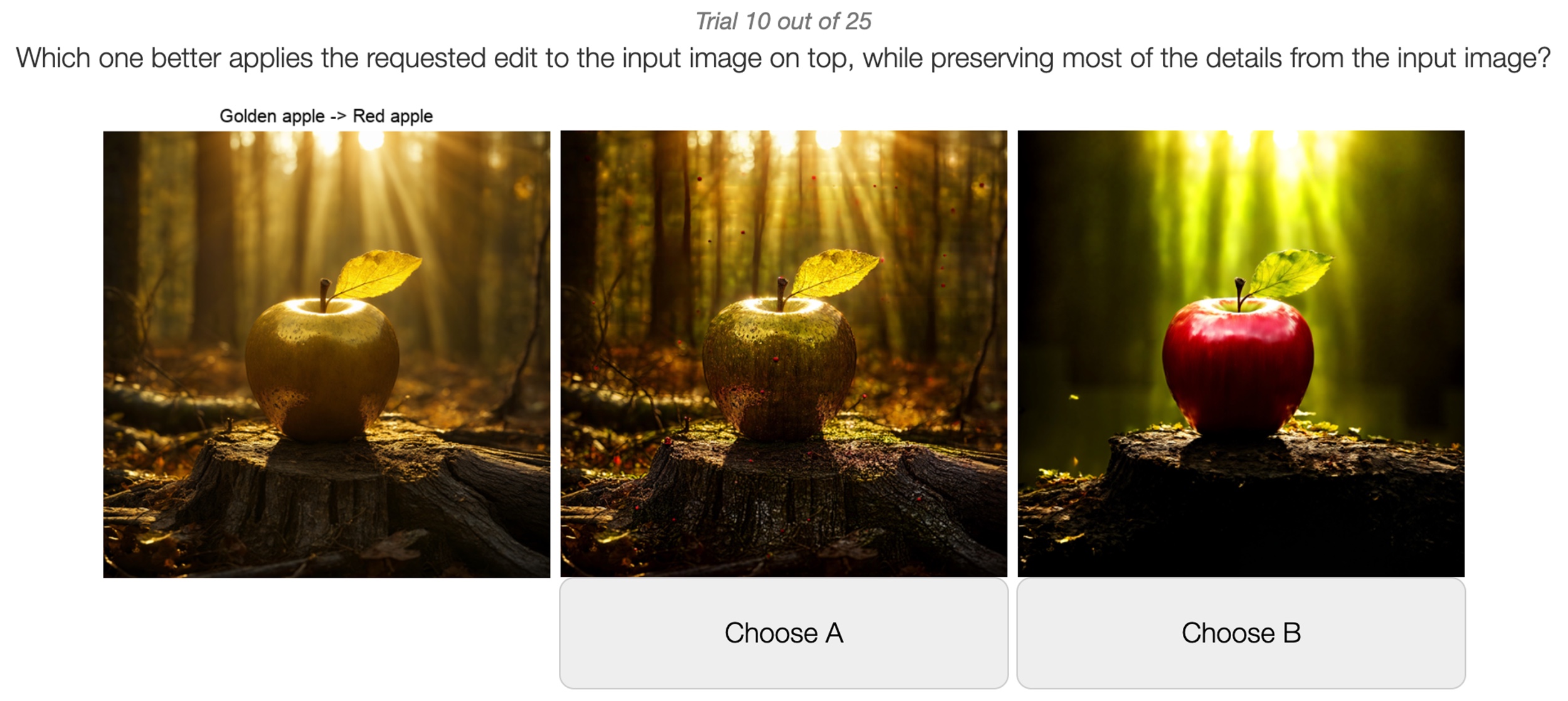}\\
    \includegraphics[width=0.8\textwidth]{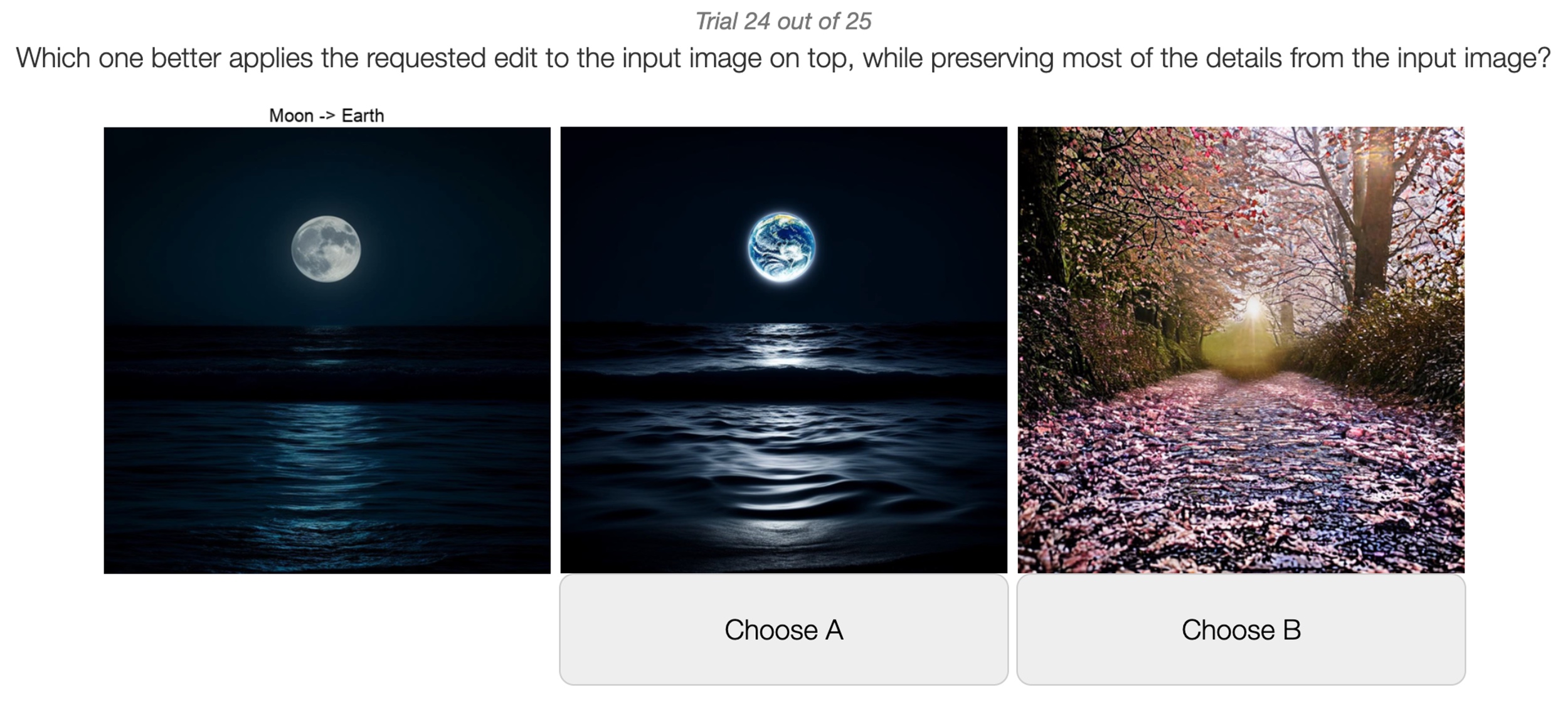}
    \caption{Screen captures of user study. The top example illustrates a main question from the user study, while the bottom example represents a vigilance question.}
    \label{fig:user_study_main}
\end{figure*}

In Sec.~\ref{sec:experiments} of the main paper, we reported the preference statistics collected from 112 user study participants who passed the vigilance tests from Amazon MTurk. We provide additional details of the user study in the following. We instructed participants to select the most anticipated outcome when the displayed source image is edited by the text prompt with the question used in \cite{Mokady:2023NTI}:
\texttt{Which one better applies the requested edit to the input image on top, while preserving most of the details from the input image?}
The example of user study screen is shown in Fig.~\ref{fig:user_study_main}.

\section{Quantitative Evaluation of Low-Resolution Editing Combined with Super-Resolution}
\label{sec:sr_comparsion}

\begin{figure*}[!ht]
    \captionsetup{type=figure, labelfont=bf}
    \centering
    \includegraphics[width=1.0\textwidth]{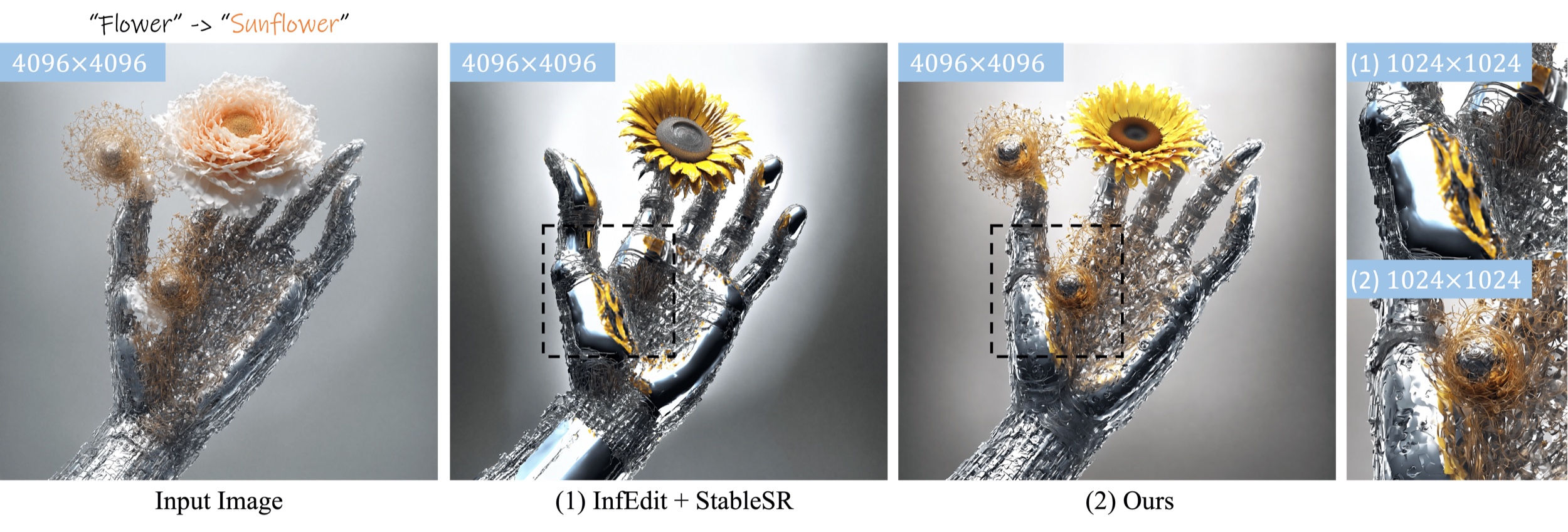}
    \caption{InfEdit+StableSR vs. EditCrafter for the teaser image.}
    \label{fig:sr_limitation}
\end{figure*}

To the best of our knowledge, apart from CSD~\cite{Kim:2023CSD}, no existing work directly addresses high-resolution image editing. However, for a comprehensive evaluation, we present quantitative comparisons in Tab.~\ref{tab:sr_quantitative_results} on our dataset against ProxEdit~\cite{Han:2024ProxEdit}+StableSR~\cite{wang2024stablesr} and InfEdit~\cite{Xu:2024InfEdit}+StableSR~\cite{wang2024stablesr}, which are currently state-of-the-art image editing methods. 
Our method, \projname{}, achieves the highest scores in both the ImageReward~\cite{xu2023imagereward} and CLIPScore~\cite{hessel2021clipscore} metrics. Furthermore, although InfEdit + StableSR attains high HPSv2 scores, it is unable to capture intricate details because resizing disrupts high-level information and subsequent super-resolution fails to recover these details, as demonstrated in Fig.~\ref{fig:sr_limitation}.

Furthermore, we conducted two user studies to compare our method against InfEdit + StableSR and ProxEdit + StableSR, respectively, using Amazon MTurk, following the same setup described in Sec.~\ref{sec:user_study}. 
We collected a total of 25 responses, including 5 vigilance tasks, from 124 participants for the comparison between our method and InfEdit + StableSR, and from 117 participants for the comparison with ProxEdit + StableSR. The results demonstrate that human evaluators preferred our \projname{} method in \textbf{61.12\%}, and \textbf{92.38\%} of cases when compared to InfEdit + StableSR, and ProxEdit + StableSR, respectively. These results demonstrate that \projname{} more effectively applies the intended edits, achieving better alignment with user expectations compared to low-resolution editing combined with super-resolution.


\begin{table}[!htbp]
\captionsetup{type=table, labelfont=bf}
    \centering
    \begin{tabular}{cc|cc|cc}
    \toprule
    CSD & Ours & InfEdit + StableSR & Ours & ProxEdit + StableSR & Ours \\
    \midrule
    27.39\% & \textbf{72.61\%} & 38.88\% & \textbf{61.12\%} & 6.62\% & \textbf{92.38\%} \\
    \bottomrule
    \end{tabular}
    \caption{User study results. Participants were instructed to select the most preferred editing outcome based on its fidelity to both the original input image and the given textual edit instruction.}
    \label{tab:user_study}
\end{table}

\begin{table}[!htbp]
\captionsetup{type=table, labelfont=bf}
    \centering
    \resizebox{0.5\linewidth}{!}{
    \begin{tabular}{lccccc}
    \toprule
        Res & Method & ImageReward $\uparrow$ & HPSv2 $\uparrow$ & CLIPScore $\uparrow$\\
        \midrule
        \multirow{2}{*}{4$\times$ 1:1} 
        & CSD & 0.5538 & 0.2883 & 32.8353 \\
        & InfEdit+SR & 1.2212 & \textbf{0.2982} & 33.6893 \\
        & ProxEdit+SR & -0.5561 & 0.2833 & 30.3980 \\
        & Ours & \textbf{1.4831} & 0.2935 & \textbf{34.8039} \\
        \midrule
        \multirow{2}{*}{8$\times$ 1:2} 
        & CSD & 0.7165 & 0.2782 & 32.2794 \\
        & Ours & \textbf{1.4238} & \textbf{0.2824} & \textbf{34.5303} \\
        \midrule
        \multirow{2}{*}{16$\times$ 1:1} 
        & CSD & 0.6304 & 0.2934 & 32.7795 \\
        & InfEdit+SR & 1.6670 & \textbf{0.3021} & 35.1438 \\
        & ProxEdit+SR & 0.5440 & 0.2873 & 32.6323 \\
        & Ours & \textbf{1.6689} & 0.3017 & \textbf{35.3194} \\
        \bottomrule
    \end{tabular}
    }
    \caption{Quantitative comparisons on SD2.1.
    }
    \label{tab:sr_quantitative_results}
\end{table}

\onecolumn
\clearpage
\section{High Quality Version of Fig.~\ref{fig:main_comparison}}
\label{sec:main_comparison_full}

{
\small
\setlength{\tabcolsep}{0em}
\def\arraystretch{0.0}
\begin{longtable}[h!]{
>{\centering\arraybackslash}m{0.10\textwidth}
>{\centering\arraybackslash}m{0.10\textwidth}
>{\centering\arraybackslash}m{0.10\textwidth}
>{\centering\arraybackslash}m{0.10\textwidth}
>{\centering\arraybackslash}m{0.10\textwidth}
>{\centering\arraybackslash}m{0.10\textwidth}
>{\centering\arraybackslash}m{0.10\textwidth}
>{\centering\arraybackslash}m{0.10\textwidth}
>{\centering\arraybackslash}m{0.10\textwidth}
>{\centering\arraybackslash}m{0.10\textwidth}
}
    \endhead
    \multicolumn{9}{c}{\hspace*{0cm}\raisebox{-0.5\totalheight}{\includegraphics[width=1.0\textwidth]{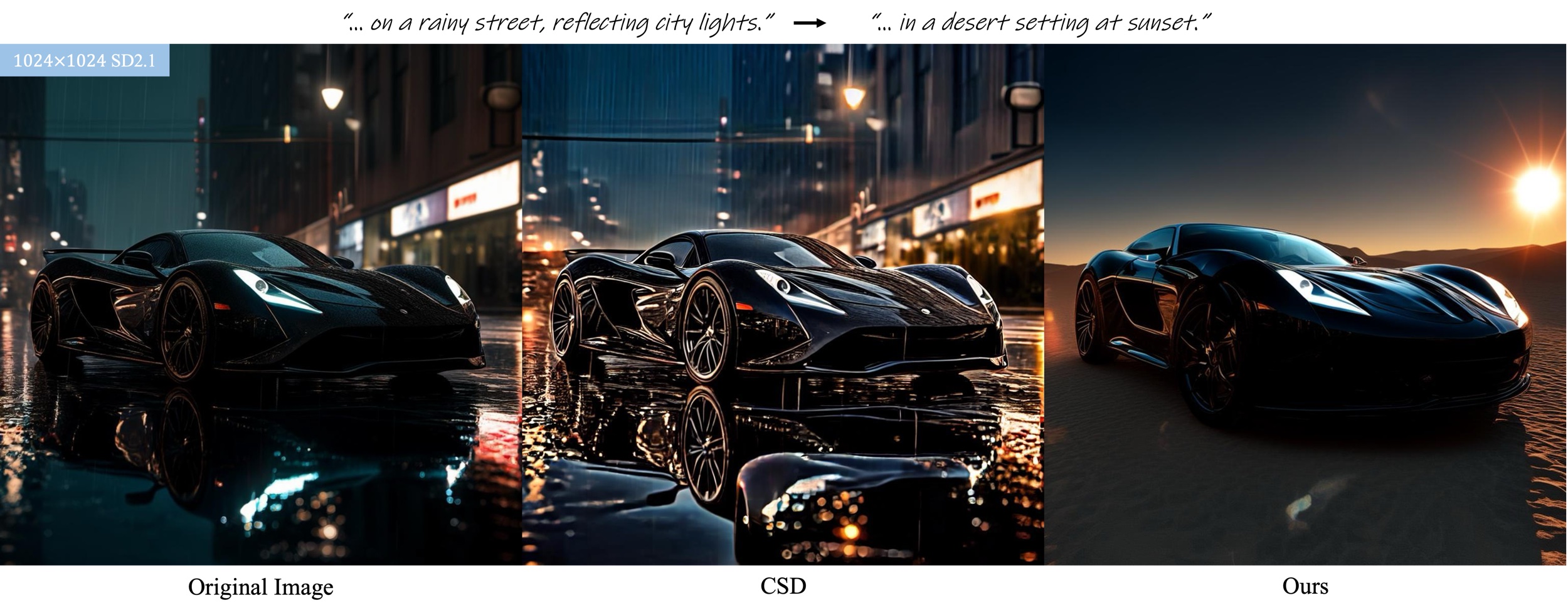}}}  \\
    \\[+0.5em]
    \multicolumn{9}{c}{\hspace*{0cm}\raisebox{-0.5\totalheight}{\includegraphics[width=1.0\textwidth]{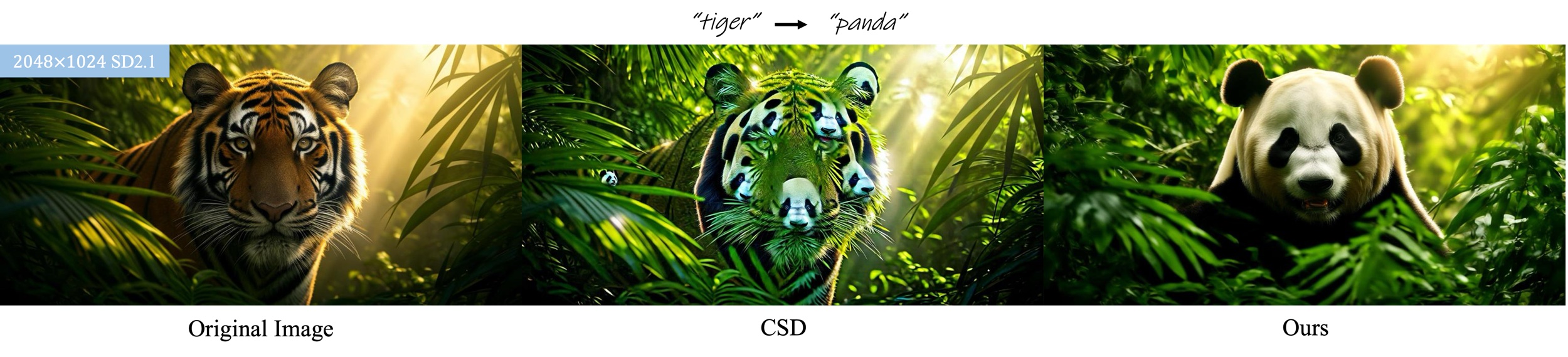}}}  \\
    \\[+0.5em]
    \multicolumn{9}{c}{\hspace*{0cm}\raisebox{-0.5\totalheight}{\includegraphics[width=1.0\textwidth]{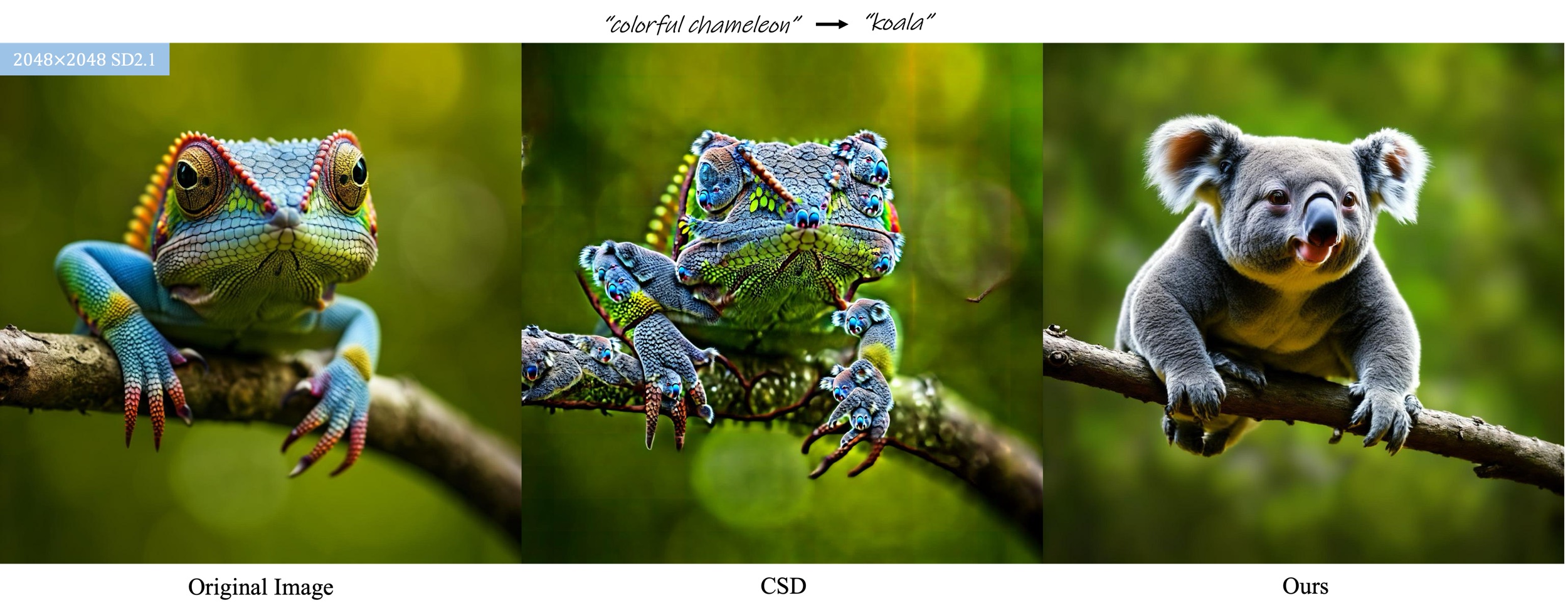}}}  \\
    \\[+0.5em]
    \multicolumn{9}{c}{\hspace*{0cm}\raisebox{-0.5\totalheight}{\includegraphics[width=1.0\textwidth]{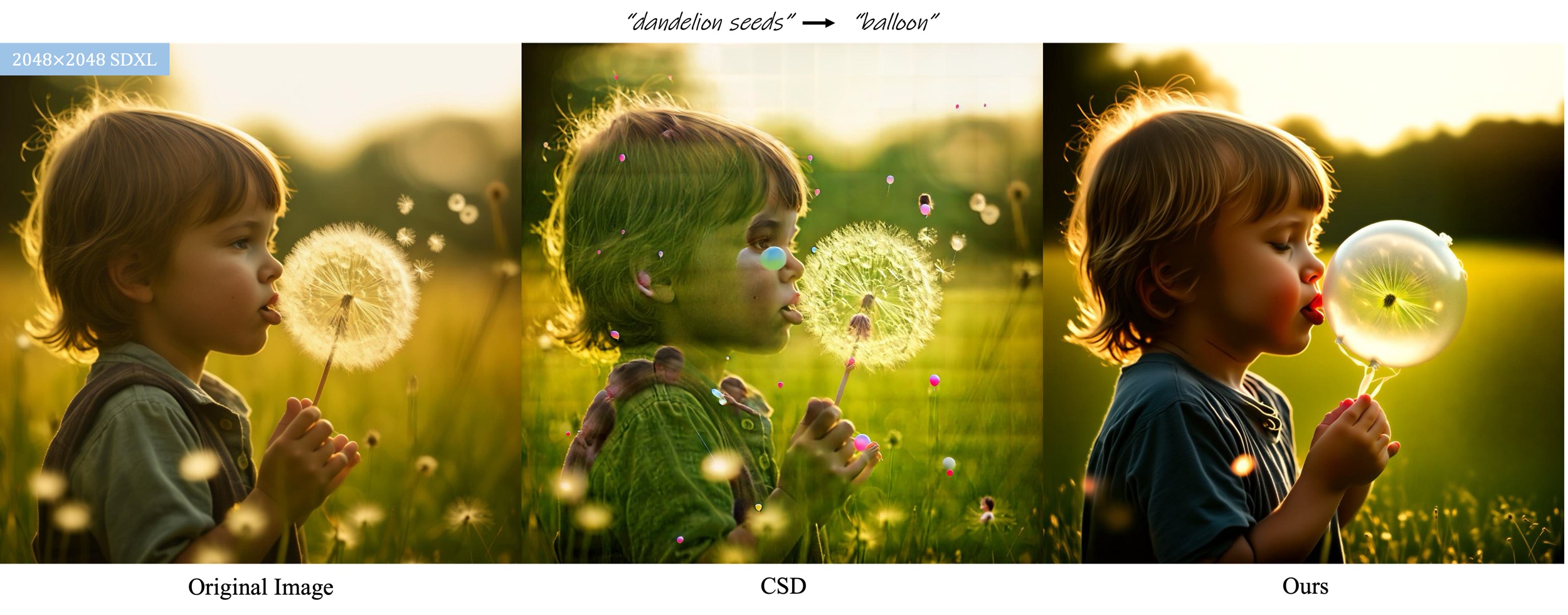}}}  \\
    \\[+0.5em]

    \scriptsize{\makecell{Original\\Image}} \vspace{0.1cm}&
    \multicolumn{9}{c}{\hspace*{0cm}\raisebox{-0.5\totalheight}{\includegraphics[width=0.895\textwidth]{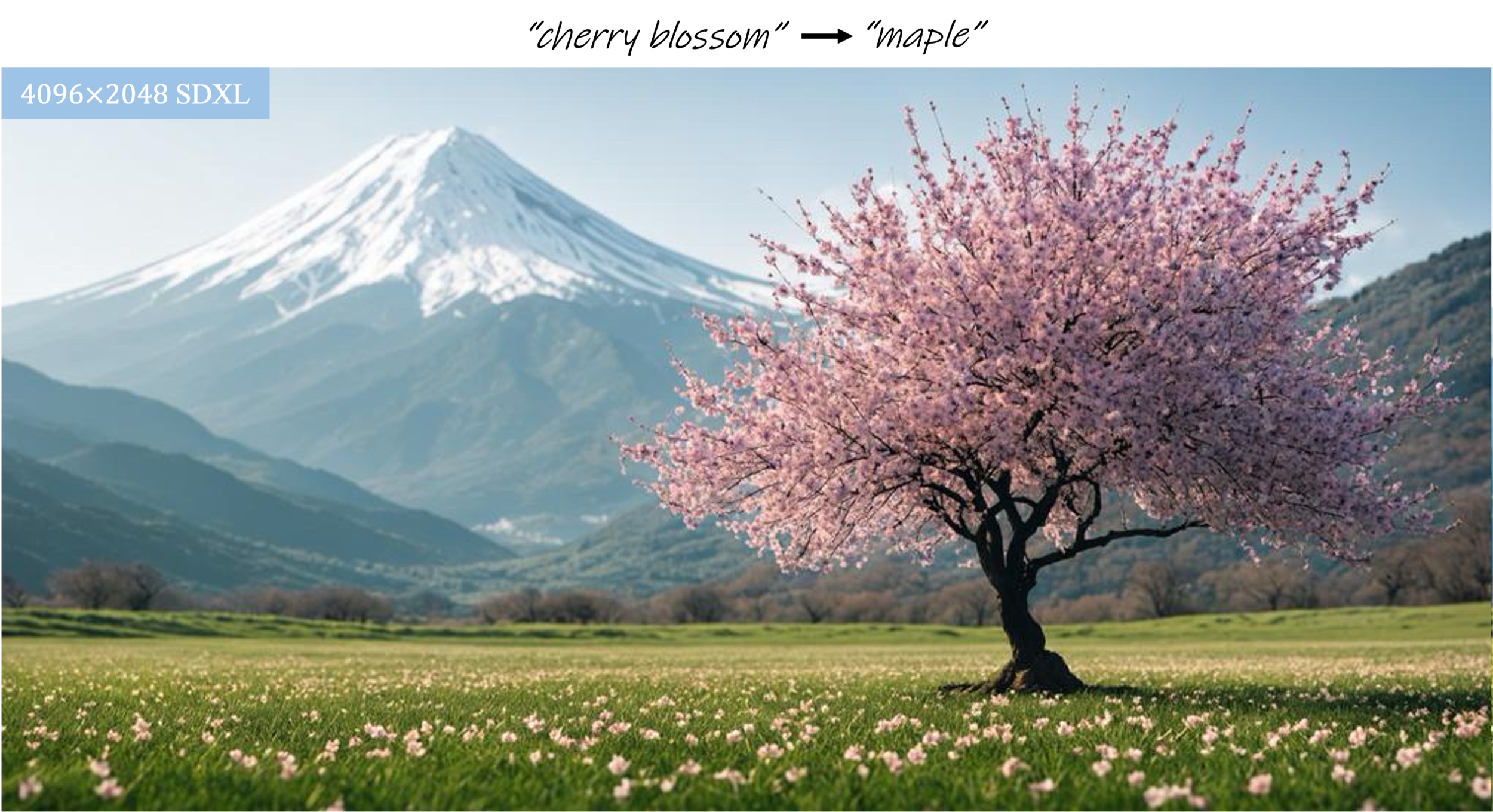}}}  \\
    \\[+0.5em]
    \scriptsize{CSD~\cite{Kim:2023CSD}} \vspace{0.1cm}&
    \multicolumn{9}{c}{\hspace*{0cm}\raisebox{-0.5\totalheight}{\includegraphics[width=0.895\textwidth]{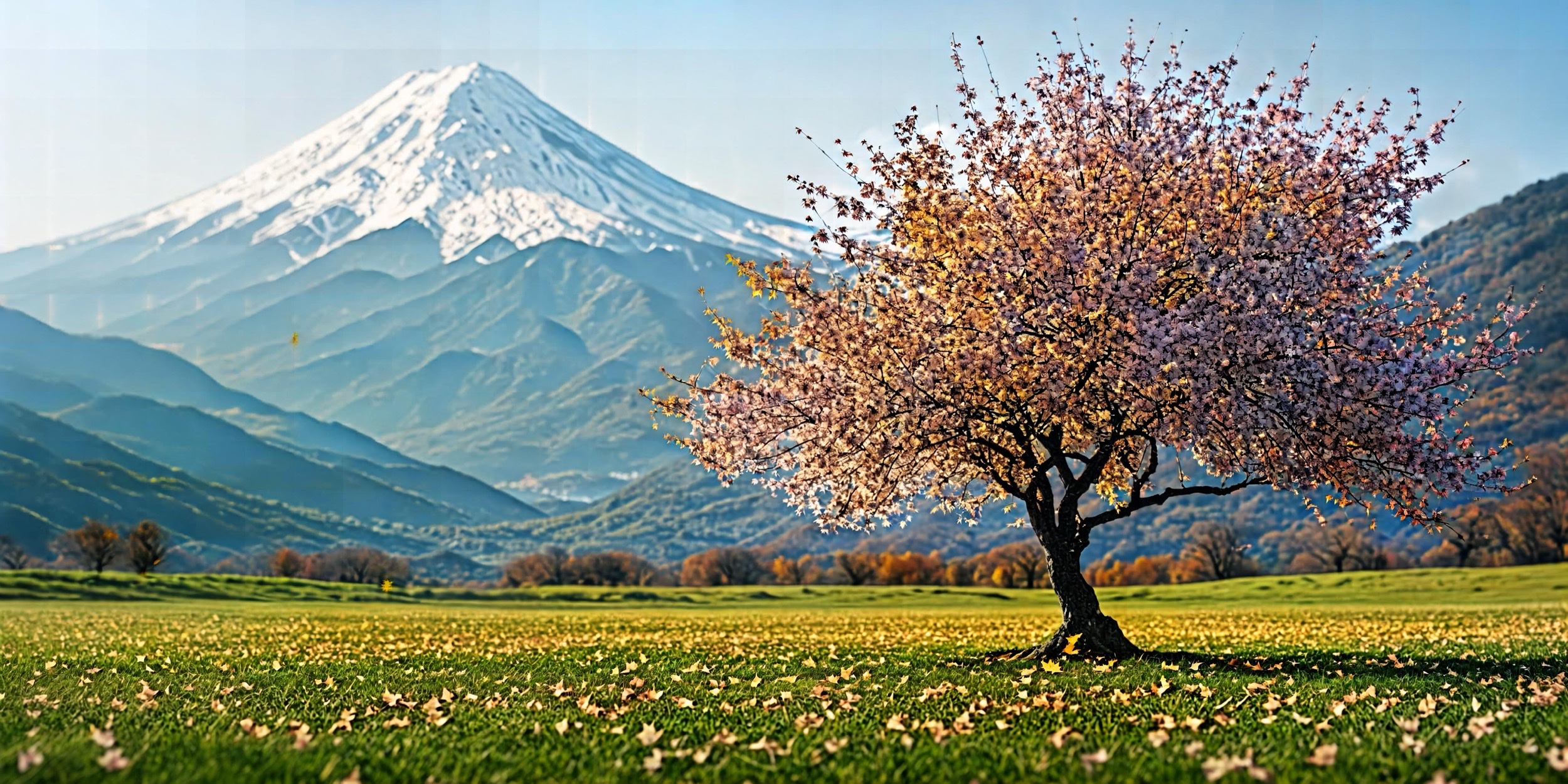}}}  \\
    \\[+0.5em]
    \scriptsize{\makecell{\textsc{Edit}\\\textsc{-Crafter}}} \vspace{0.1cm}&
    \multicolumn{9}{c}{\hspace*{0cm}\raisebox{-0.5\totalheight}{\includegraphics[width=0.895\textwidth]{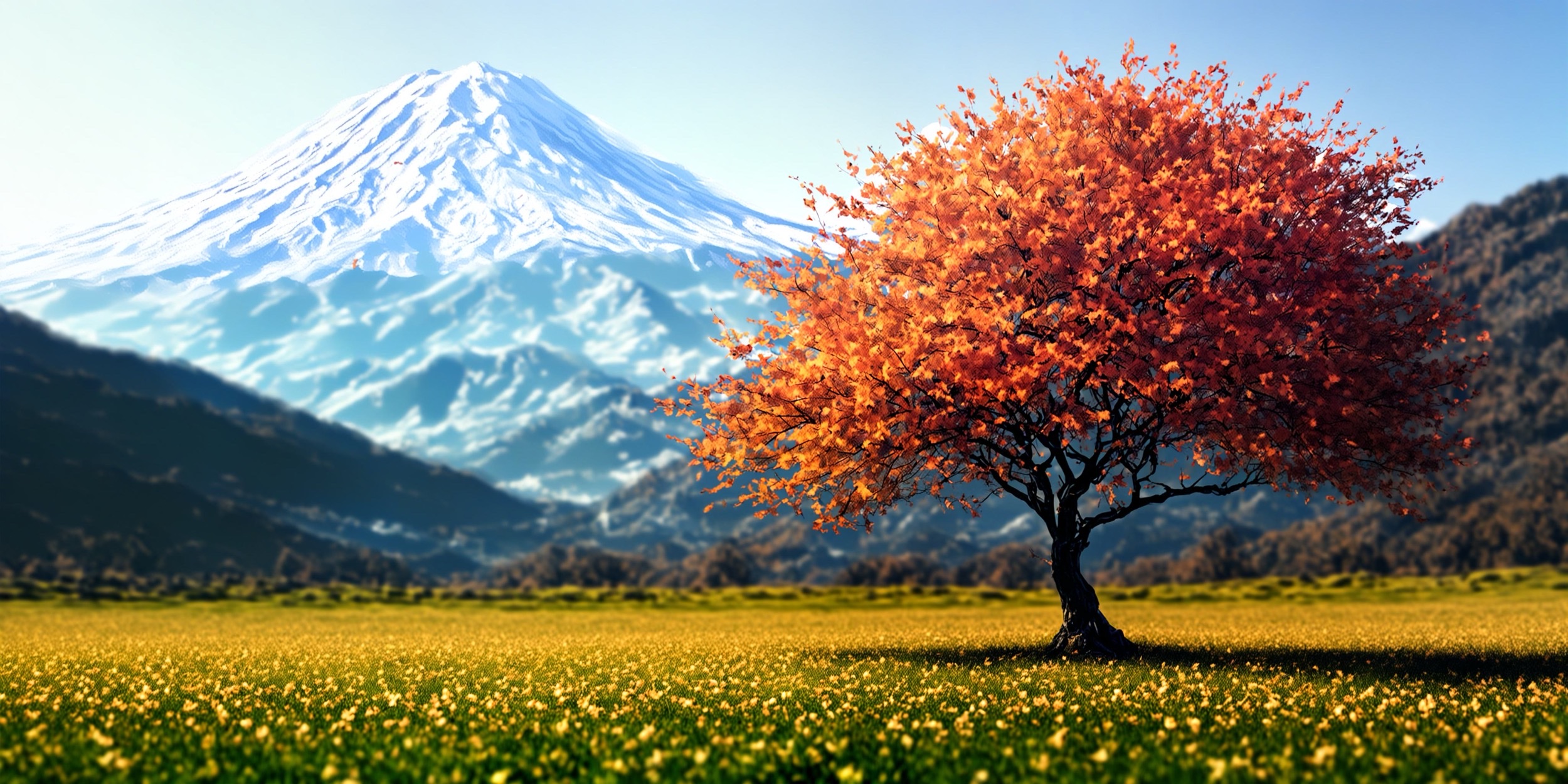}}}  \\
    \\[+0.5em]
    \pagebreak

    \scriptsize{\makecell{Original\\Image}} \vspace{0.1cm}&
    \multicolumn{9}{c}{\hspace*{0cm}\raisebox{-0.5\totalheight}{\includegraphics[width=0.895\textwidth]{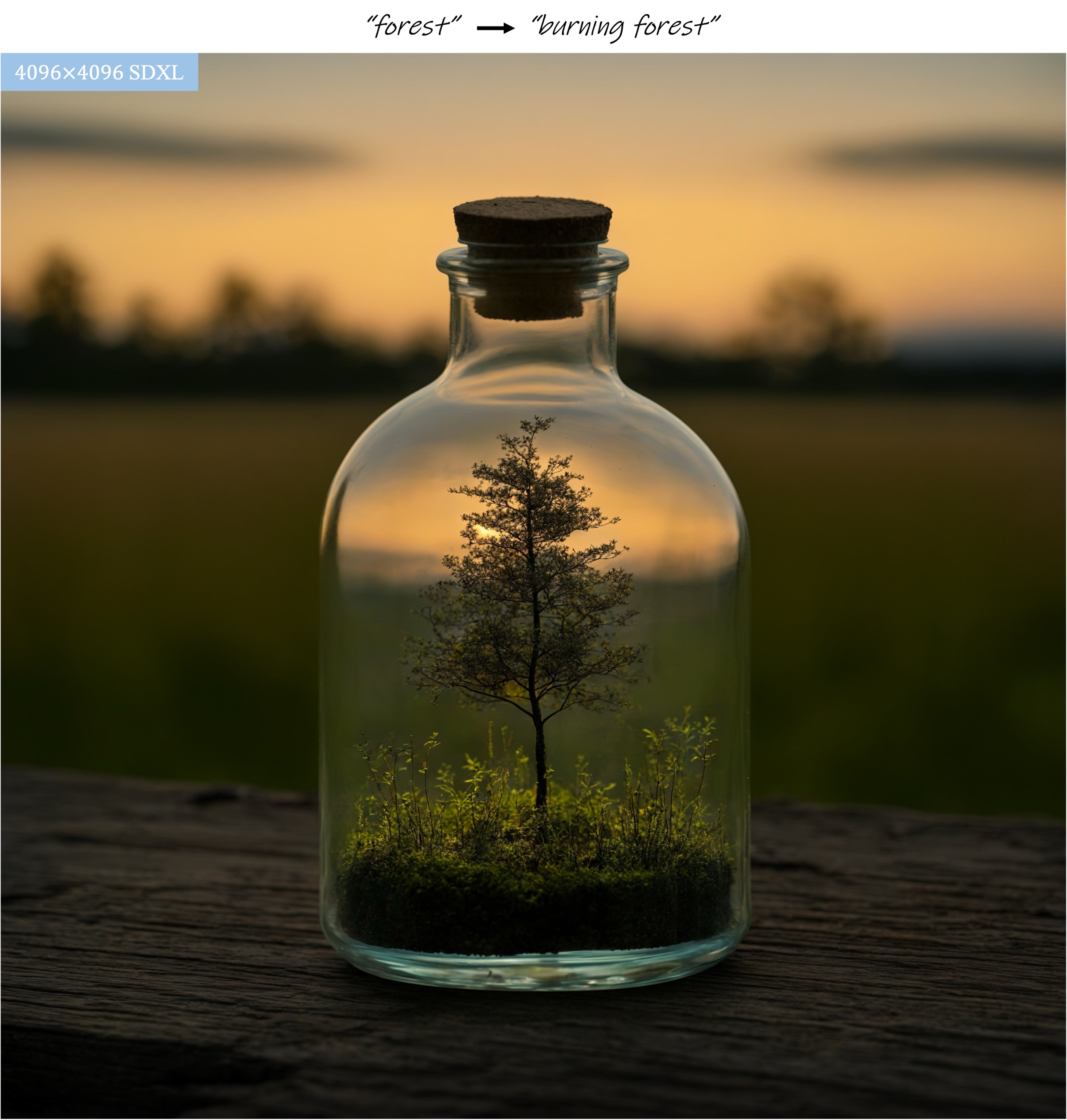}}}  \\
    \\[+0.5em]
    \scriptsize{CSD~\cite{Kim:2023CSD}} \vspace{0.1cm}&
    \multicolumn{9}{c}{\hspace*{0cm}\raisebox{-0.5\totalheight}{\includegraphics[width=0.895\textwidth]{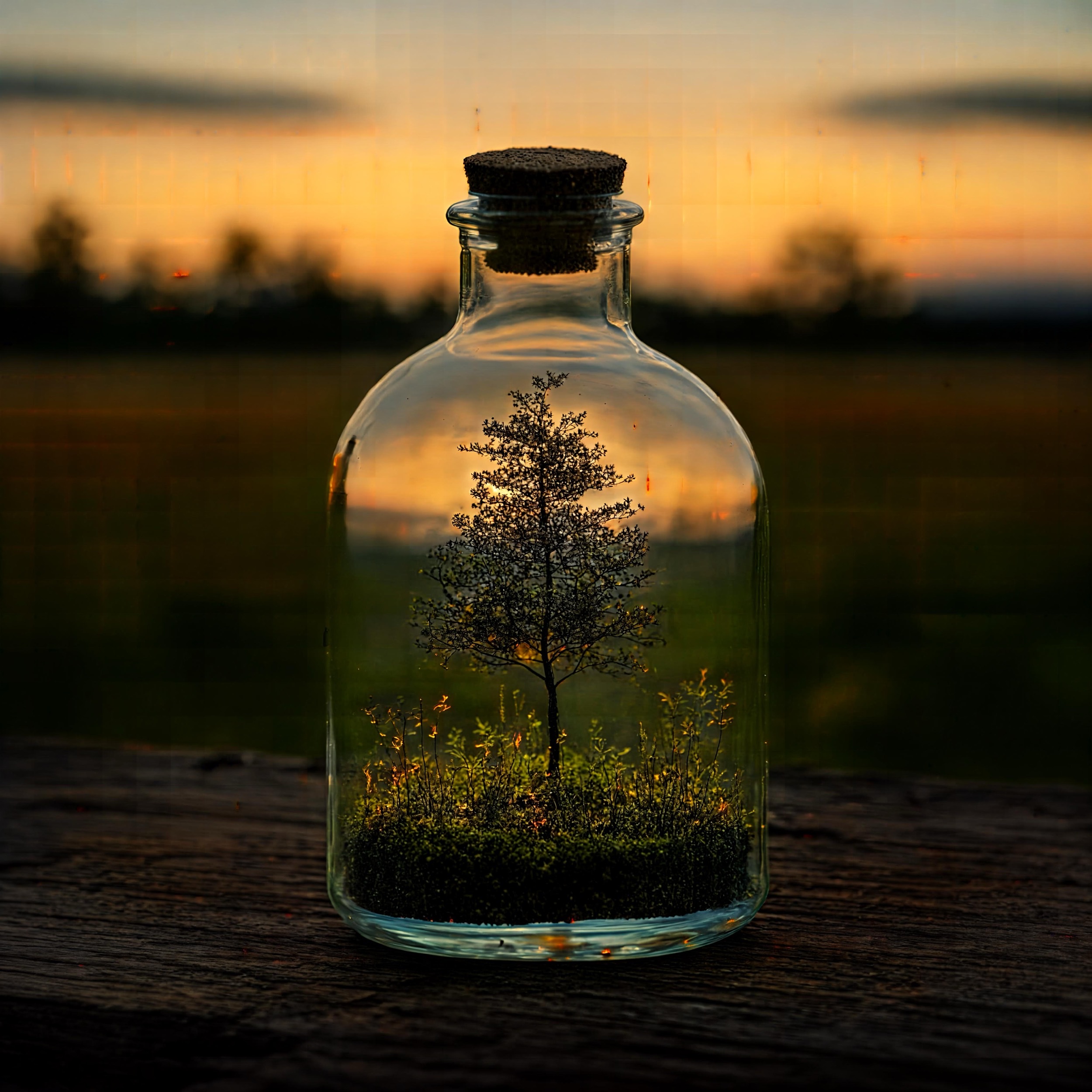}}}  \\
    \\[+0.5em]
    \scriptsize{\makecell{\textsc{Edit}\\\textsc{-Crafter}}} \vspace{0.1cm}&
    \multicolumn{9}{c}{\hspace*{0cm}\raisebox{-0.5\totalheight}{\includegraphics[width=0.895\textwidth]{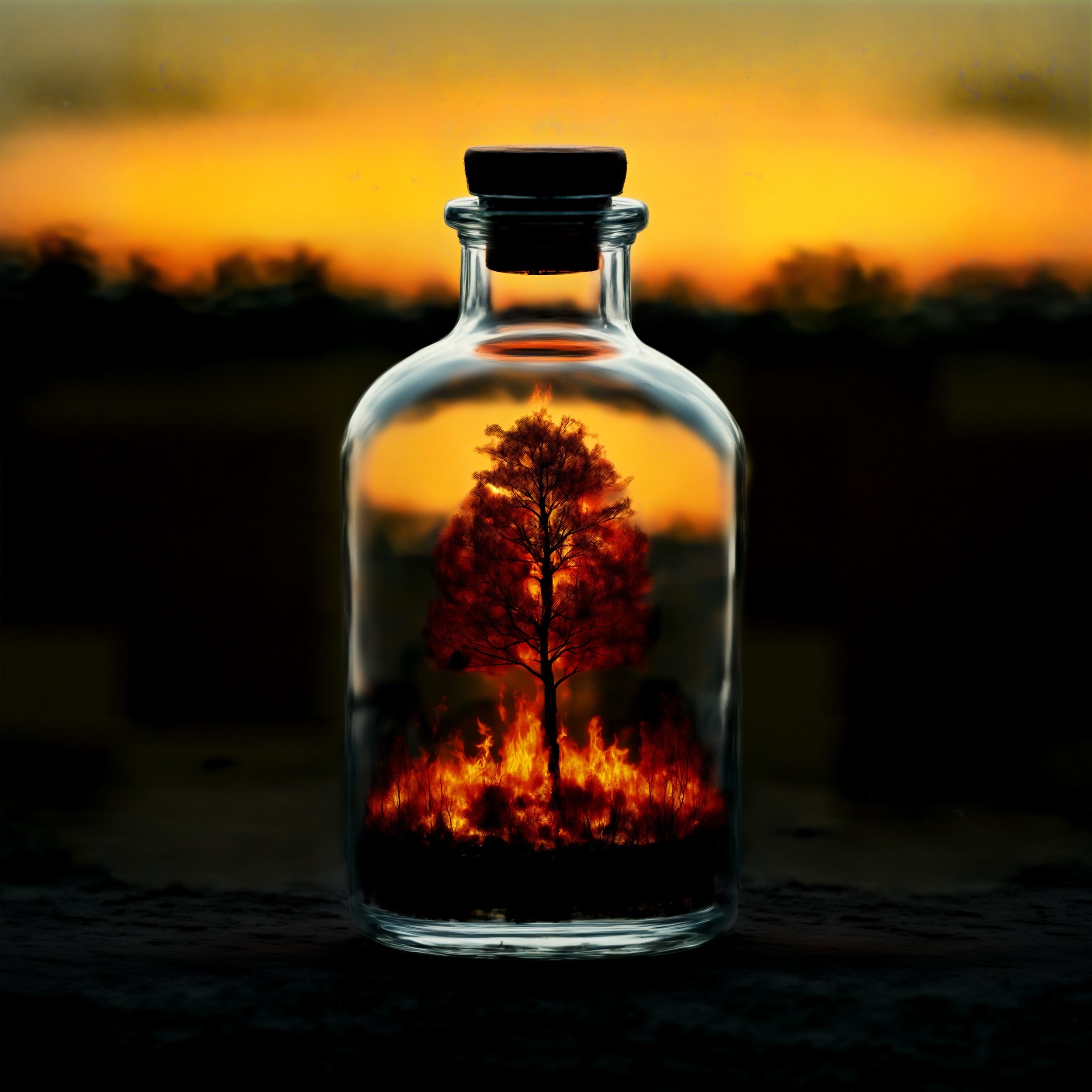}}}  \\
    \\[+0.5em]


\end{longtable}
}


\clearpage

\clearpage
\newpage
\section{More Qualitative Comparisons}
\label{sec:more_comparisons}

{
\small
\setlength{\tabcolsep}{0em}
\def\arraystretch{0.0}
\begin{longtable}[h!]{
>{\centering\arraybackslash}m{0.33\textwidth}
>{\centering\arraybackslash}m{0.33\textwidth}
>{\centering\arraybackslash}m{0.33\textwidth}
}
    Original Image &
    CSD~\cite{Kim:2023CSD} &
    \textsc{EditCrafter} (Ours) \\
    \midrule
    \endhead
    
    \multicolumn{3}{c}{\makecell{\\\hspace*{0cm} \textbf{SD2.1 $\times$4} \textit{``moon''} $\rightarrow$ \textit{``earth''}}}  \\
    \multicolumn{3}{c}{\hspace*{0cm}\raisebox{-0.5\totalheight}{\includegraphics[width=1.0\textwidth]{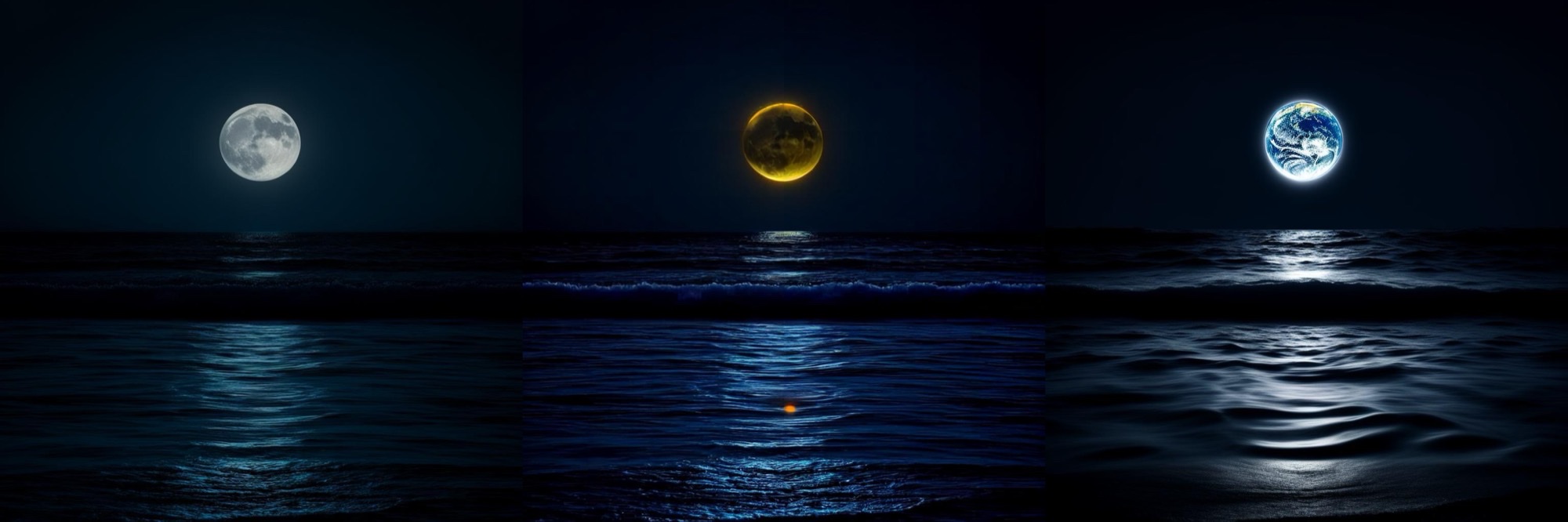}}}  \\
    \\[+0.5em]

    \multicolumn{3}{c}{\makecell{\\\hspace*{0cm} \textbf{SD2.1 $\times$4} \textit{``blanket''} $\rightarrow$ \textit{``grass''}}}  \\
    \multicolumn{3}{c}{\hspace*{0cm}\raisebox{-0.5\totalheight}{\includegraphics[width=1.0\textwidth]{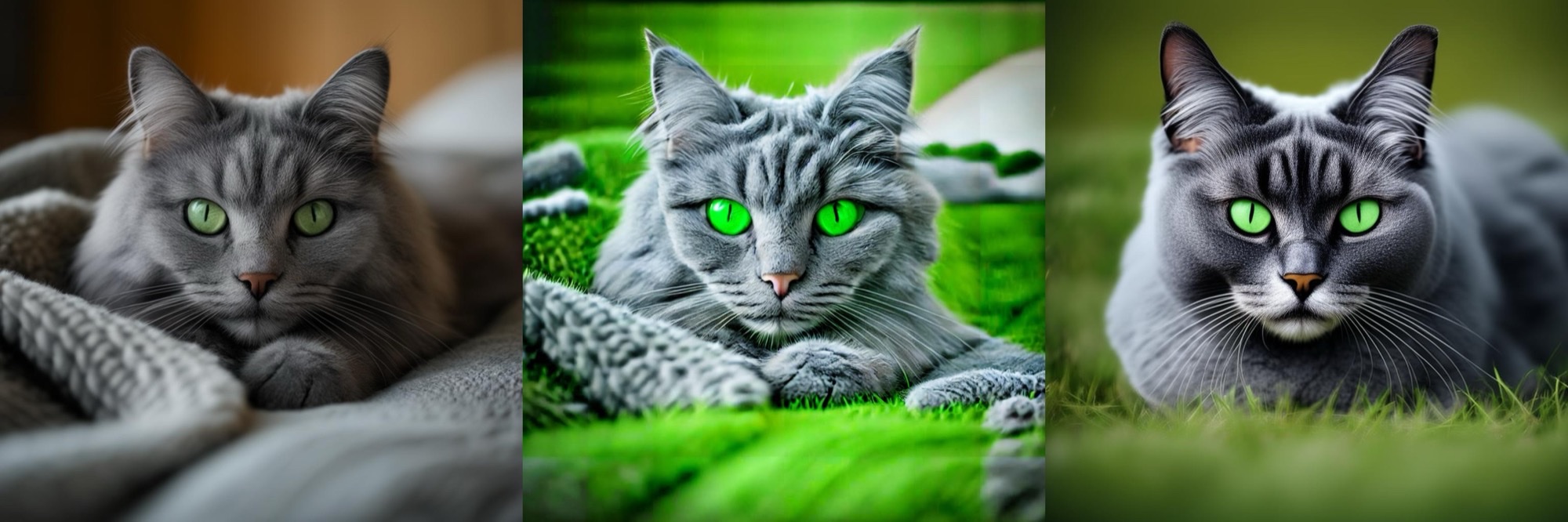}}}  \\
    \\[+0.5em]

    \multicolumn{3}{c}{\makecell{\\\hspace*{0cm} \textbf{SD2.1 $\times$4} \textit{``cactus''} $\rightarrow$ \textit{``aloe''}}}  \\
    \multicolumn{3}{c}{\hspace*{0cm}\raisebox{-0.5\totalheight}{\includegraphics[width=1.0\textwidth]{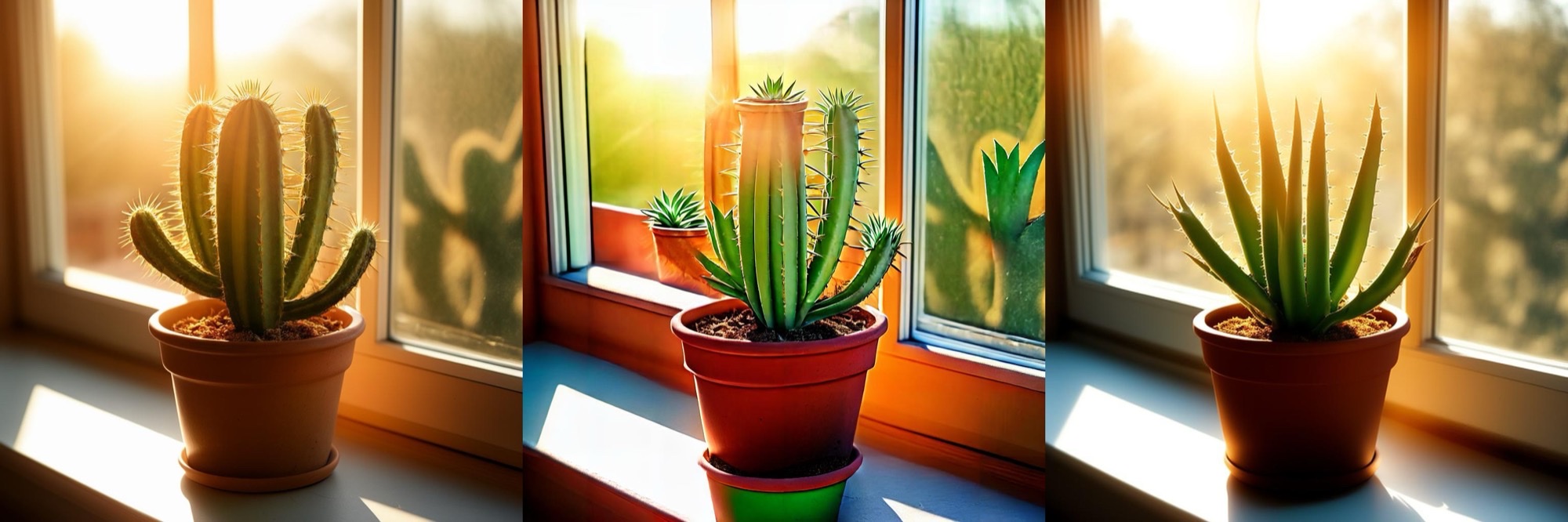}}}  \\
    \\[+0.5em]

    \newpage
    
    \multicolumn{3}{c}{\makecell{\\\hspace*{0cm} \textbf{SD2.1 $\times$4} \textit{``lemon''} $\rightarrow$ \textit{``cucumber''}}}  \\
    \multicolumn{3}{c}{\hspace*{0cm}\raisebox{-0.5\totalheight}{\includegraphics[width=1.0\textwidth]{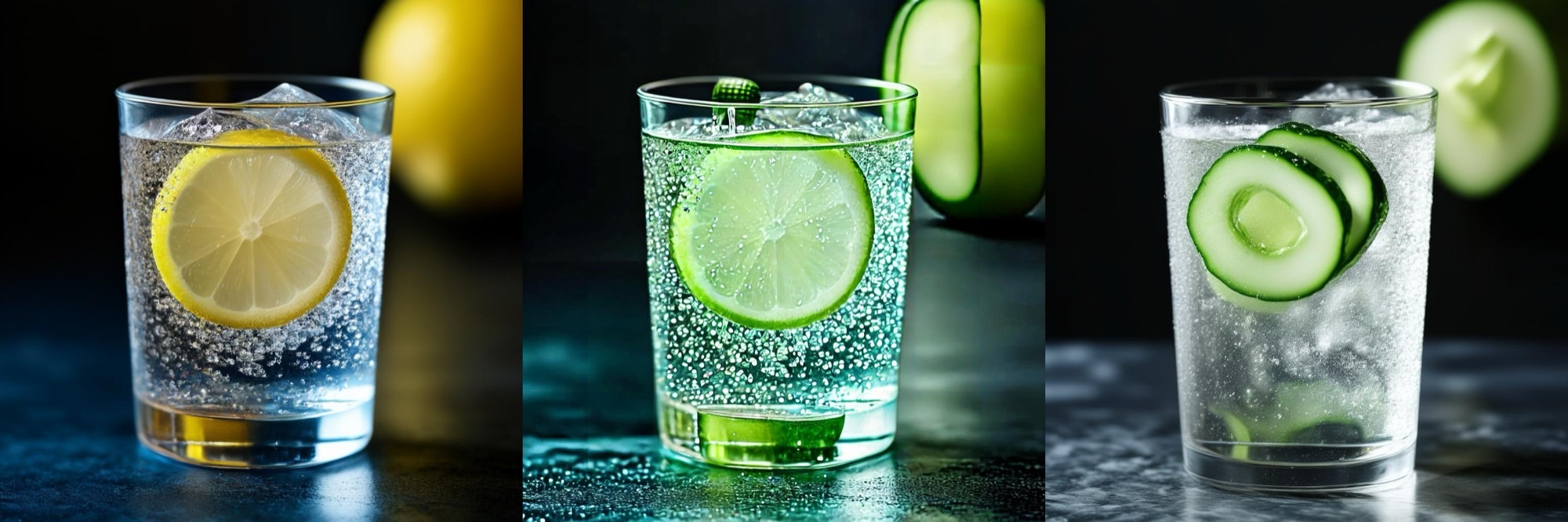}}}  \\
    \\[+0.5em]

    \multicolumn{3}{c}{\makecell{\\\hspace*{0cm} \textbf{SD2.1 $\times$4} \textit{``tulips''} $\rightarrow$ \textit{``roses''}}}  \\
    \multicolumn{3}{c}{\hspace*{0cm}\raisebox{-0.5\totalheight}{\includegraphics[width=1.0\textwidth]{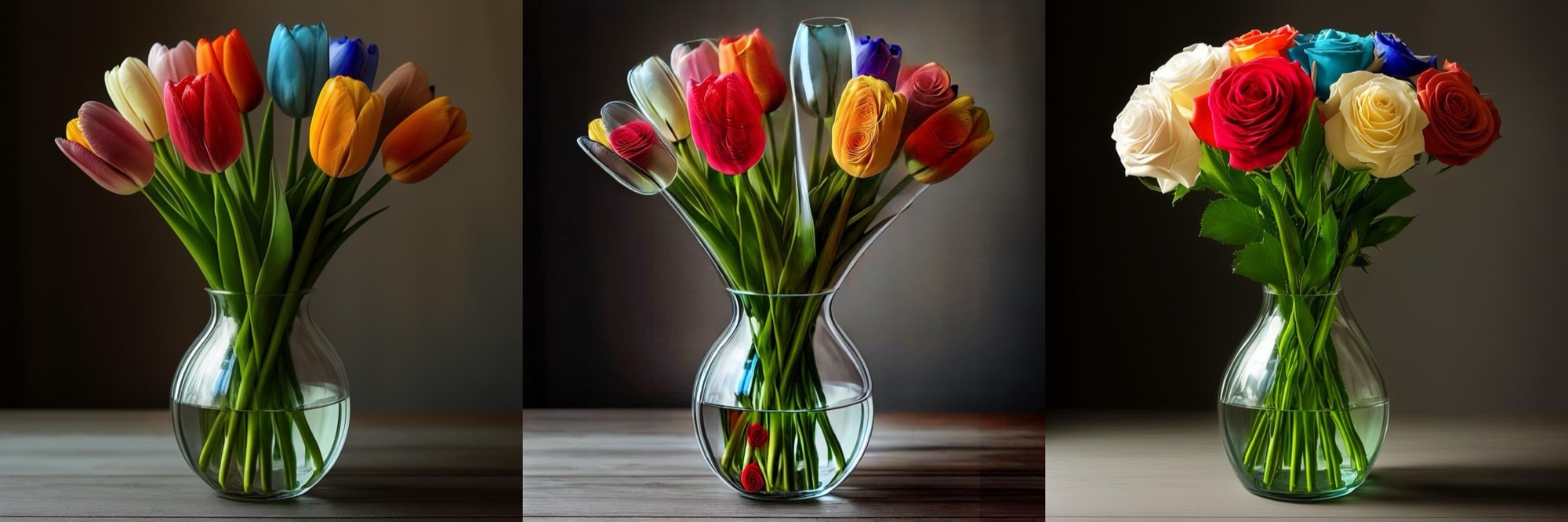}}}  \\
    \\[+0.5em]

    \multicolumn{3}{c}{\makecell{\\\hspace*{0cm} \textbf{SD2.1 $\times$4} \textit{``vilage''} $\rightarrow$ \textit{``castle''}}}  \\
    \multicolumn{3}{c}{\hspace*{0cm}\raisebox{-0.5\totalheight}{\includegraphics[width=1.0\textwidth]{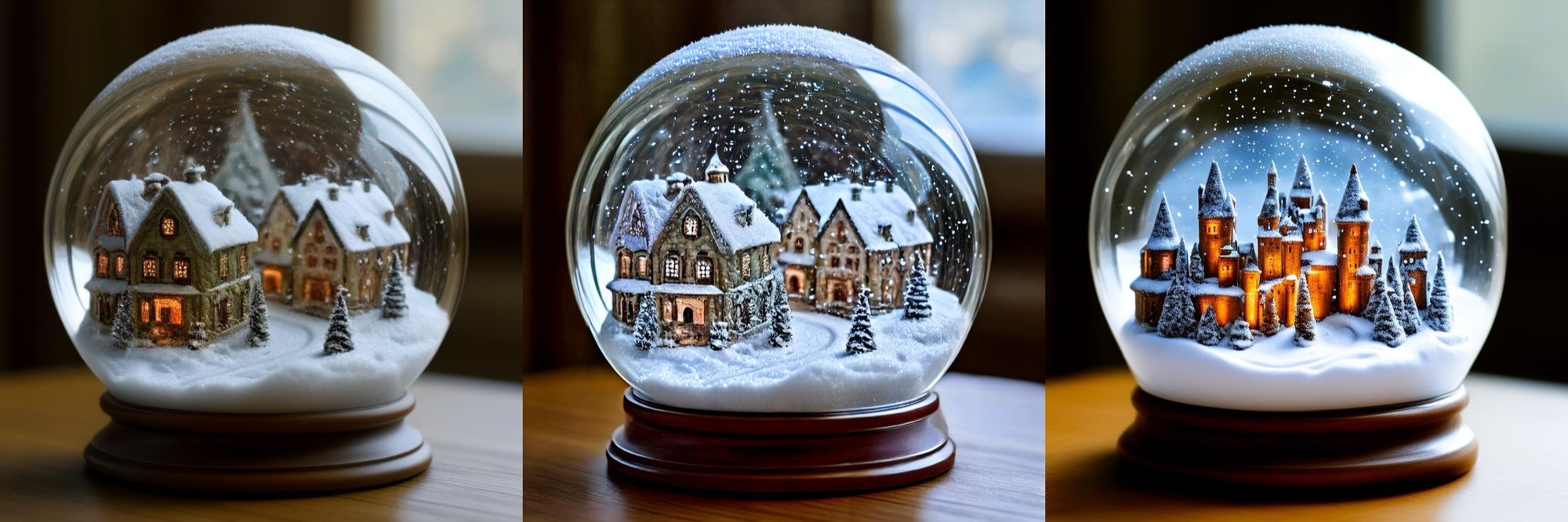}}}  \\

    \newpage

    \multicolumn{3}{c}{\makecell{\\\hspace*{0cm} \textbf{SD2.1 $\times$8} \textit{``vilage''} $\rightarrow$ \textit{``castle''}}}  \\
    \multicolumn{3}{c}{\hspace*{0cm}\raisebox{-0.5\totalheight}{\includegraphics[width=1.0\textwidth]{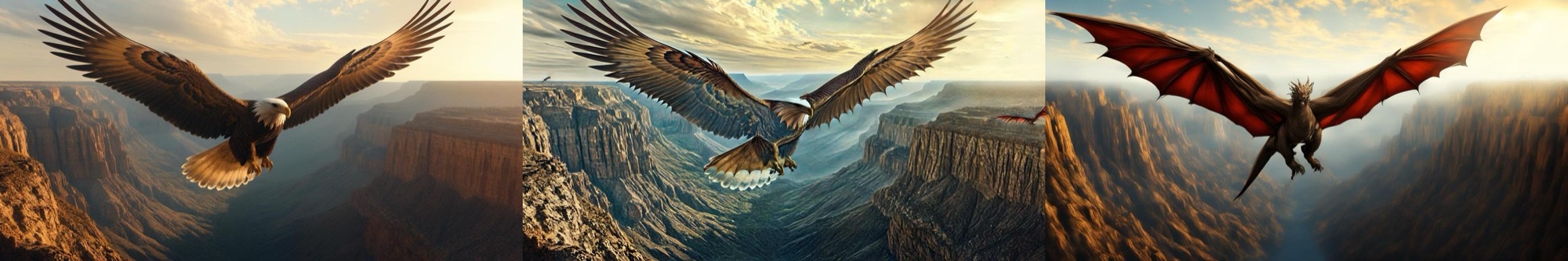}}}  \\
    \\[+0.5em]

    \multicolumn{3}{c}{\makecell{\\\hspace*{0cm} \textbf{SD2.1 $\times$8} \textit{``fox''} $\rightarrow$ \textit{``lion''}}}  \\
    \multicolumn{3}{c}{\hspace*{0cm}\raisebox{-0.5\totalheight}{\includegraphics[width=1.0\textwidth]{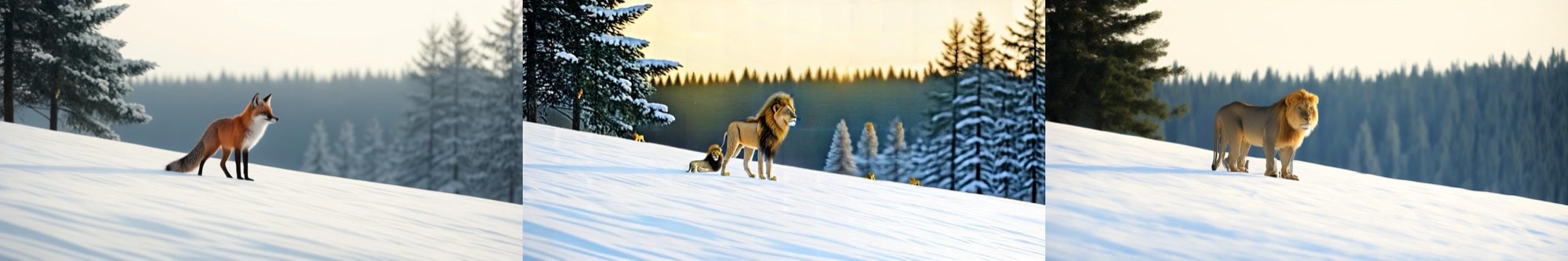}}}  \\
    \\[+0.5em]


    \multicolumn{3}{c}{\makecell{\\\hspace*{0cm} \textbf{SD2.1 $\times$8} \textit{``owl''} $\rightarrow$ \textit{``hawk''}}}  \\
    \multicolumn{3}{c}{\hspace*{0cm}\raisebox{-0.5\totalheight}{\includegraphics[width=1.0\textwidth]{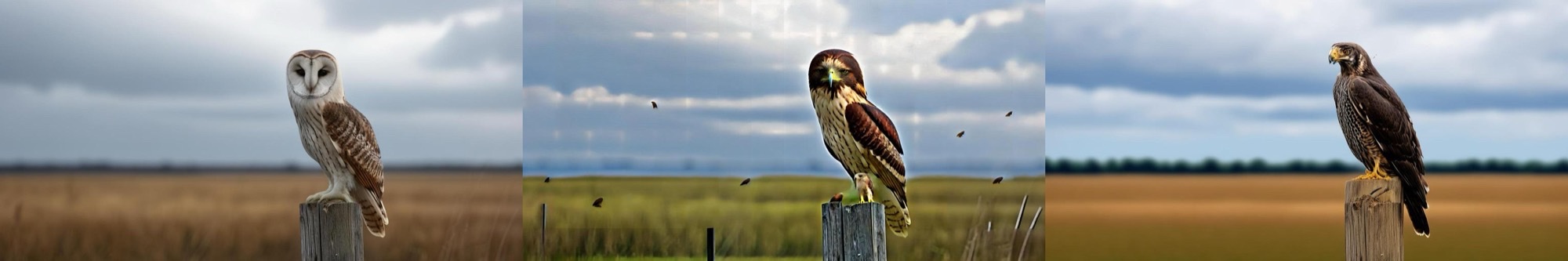}}}  \\
    \\[+0.5em]

    \multicolumn{3}{c}{\makecell{\\\hspace*{0cm} \textbf{SD2.1 $\times$8} \textit{``parm tree''} $\rightarrow$ \textit{``umbrella''}}}  \\
    \multicolumn{3}{c}{\hspace*{0cm}\raisebox{-0.5\totalheight}{\includegraphics[width=1.0\textwidth]{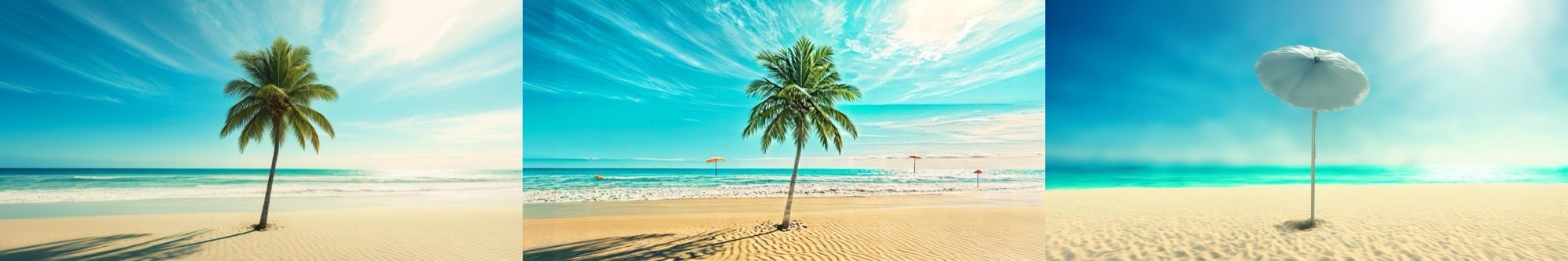}}}  \\
    \\[+0.5em]

    \multicolumn{3}{c}{\makecell{\\\hspace*{0cm} \textbf{SD2.1 $\times$8} \textit{``shark''} $\rightarrow$ \textit{``dolphin''}}}  \\
    \multicolumn{3}{c}{\hspace*{0cm}\raisebox{-0.5\totalheight}{\includegraphics[width=1.0\textwidth]{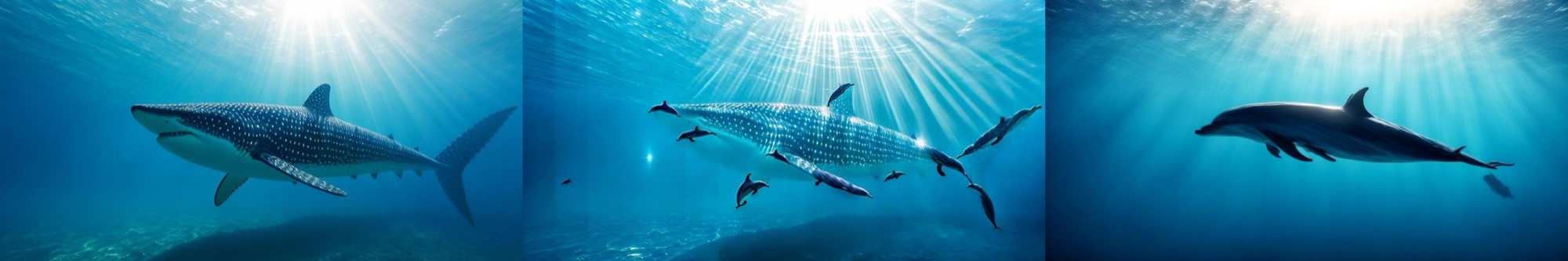}}}  \\
    \\[+0.5em]

    \newpage

    \multicolumn{3}{c}{\makecell{\\\hspace*{0cm} \textbf{SD2.1 $\times$16} \textit{``berrys''} $\rightarrow$ \textit{``roses''}}}  \\
    \multicolumn{3}{c}{\hspace*{0cm}\raisebox{-0.5\totalheight}{\includegraphics[width=1.0\textwidth]{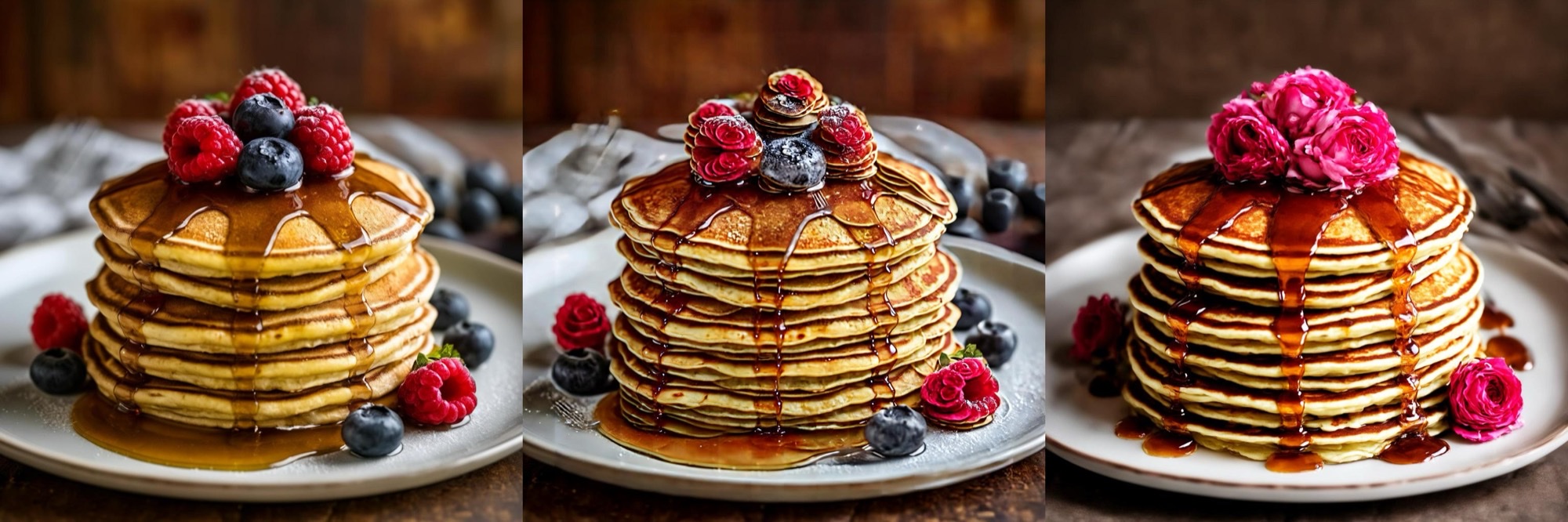}}}  \\
    \\[+0.5em]

    \multicolumn{3}{c}{\makecell{\\\hspace*{0cm} \textbf{SD2.1 $\times$16} \textit{``cat''} $\rightarrow$ \textit{``goat''}}}  \\
    \multicolumn{3}{c}{\hspace*{0cm}\raisebox{-0.5\totalheight}{\includegraphics[width=1.0\textwidth]{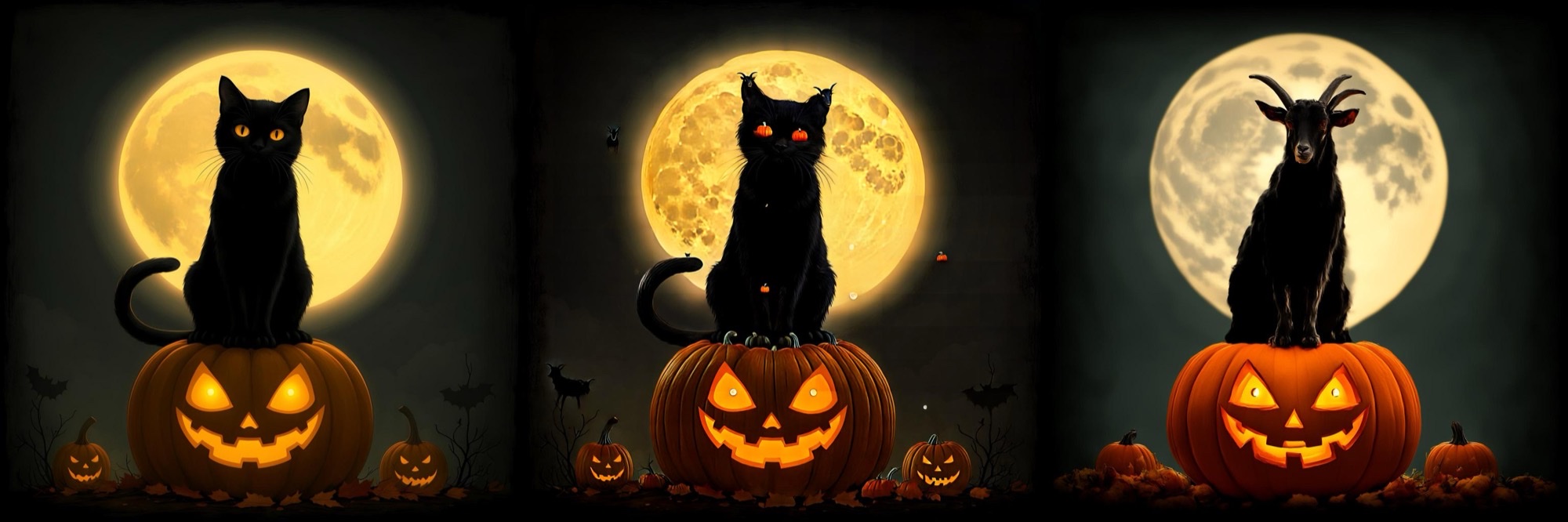}}}  \\
    \\[+0.5em]

    \multicolumn{3}{c}{\makecell{\\\hspace*{0cm} \textbf{SD2.1 $\times$16} \textit{``soccer ball''} $\rightarrow$ \textit{``crystal ball''}}}  \\
    \multicolumn{3}{c}{\hspace*{0cm}\raisebox{-0.5\totalheight}{\includegraphics[width=1.0\textwidth]{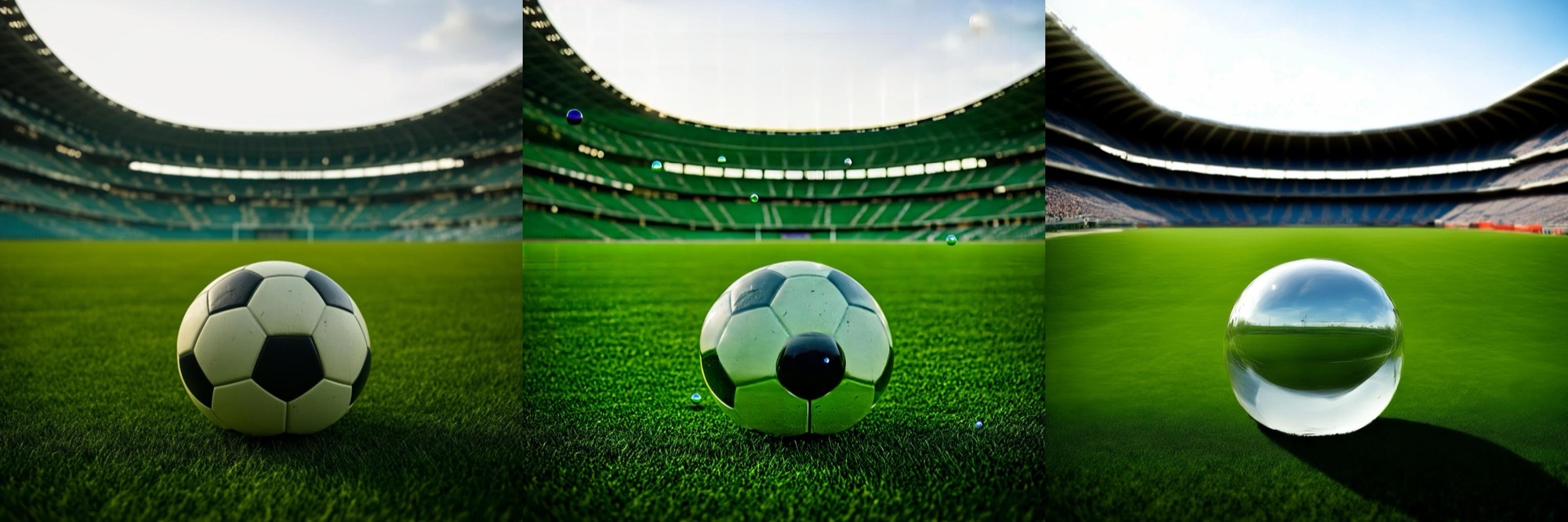}}}  \\
    \\[+0.5em]

    \newpage

    \multicolumn{3}{c}{\makecell{\\\hspace*{0cm} \textbf{SDXL $\times$4} \textit{``asphalt''} $\rightarrow$ \textit{``dessert''}}}  \\
    \multicolumn{3}{c}{\hspace*{0cm}\raisebox{-0.5\totalheight}{\includegraphics[width=1.0\textwidth]{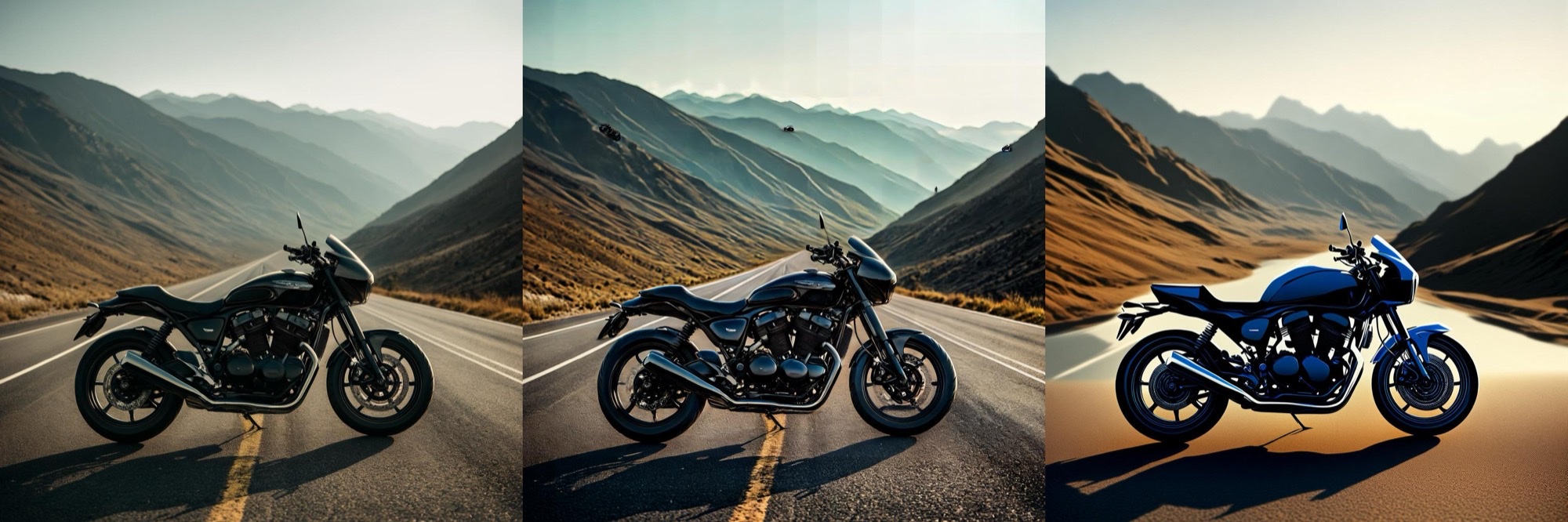}}}  \\
    \\[+0.5em]

    \multicolumn{3}{c}{\makecell{\\\hspace*{0cm} \textbf{SDXL $\times$4} \textit{``gems''} $\rightarrow$ \textit{``bones''}}}  \\
    \multicolumn{3}{c}{\hspace*{0cm}\raisebox{-0.5\totalheight}{\includegraphics[width=1.0\textwidth]{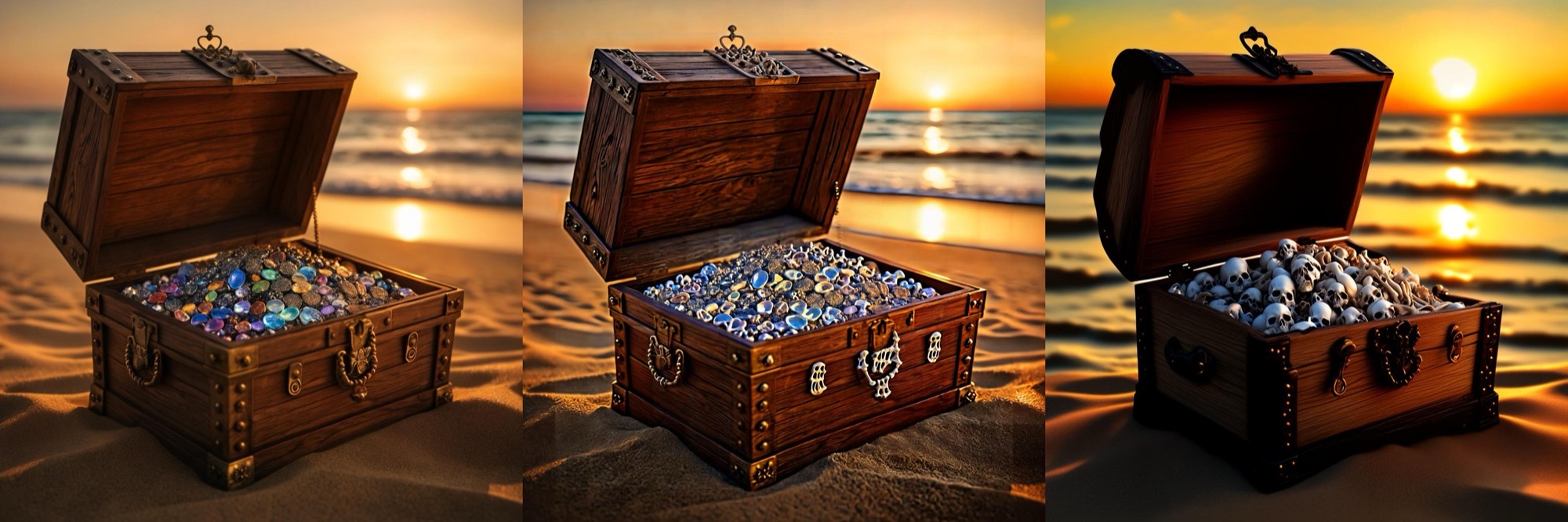}}}  \\
    \\[+0.5em] 

    \multicolumn{3}{c}{\makecell{\\\hspace*{0cm} \textbf{SDXL $\times$4} \textit{``phoenix''} $\rightarrow$ \textit{``chicken''}}}  \\
    \multicolumn{3}{c}{\hspace*{0cm}\raisebox{-0.5\totalheight}{\includegraphics[width=1.0\textwidth]{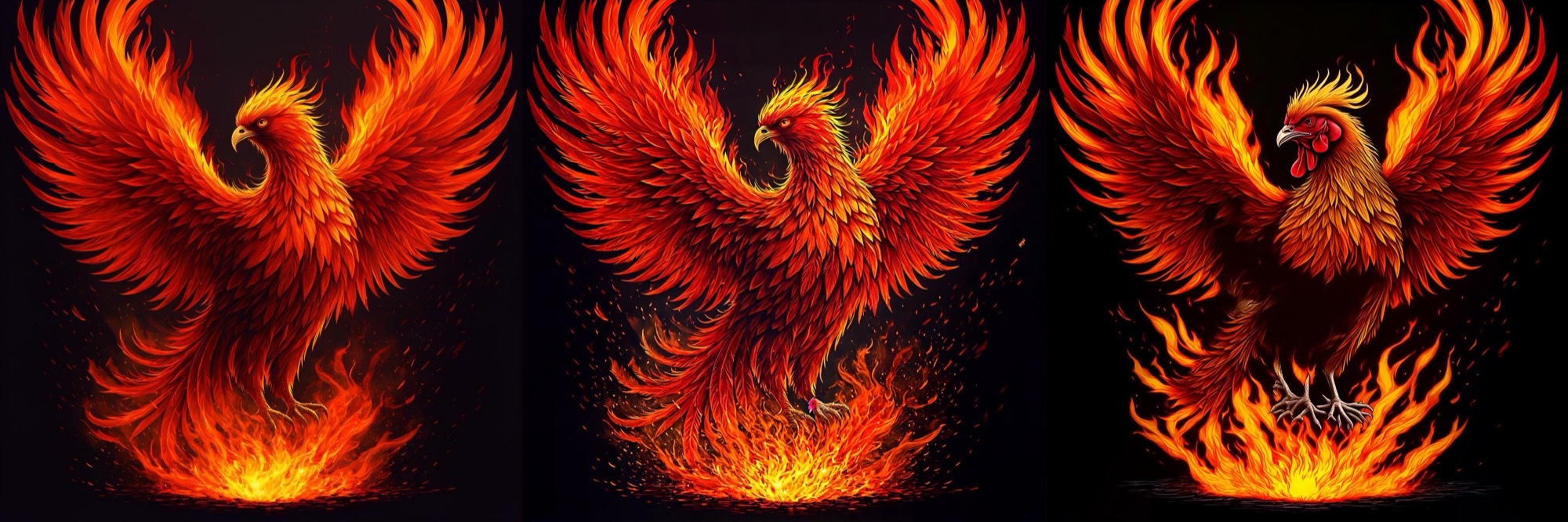}}}  \\
    \\[+0.5em] 
    
    \newpage

    \multicolumn{3}{c}{\makecell{\\\hspace*{0cm} \textbf{SDXL $\times$8} \textit{``cloud''} $\rightarrow$ \textit{``mushroom''}}}  \\
    \multicolumn{3}{c}{\hspace*{0cm}\raisebox{-0.5\totalheight}{\includegraphics[width=1.0\textwidth]{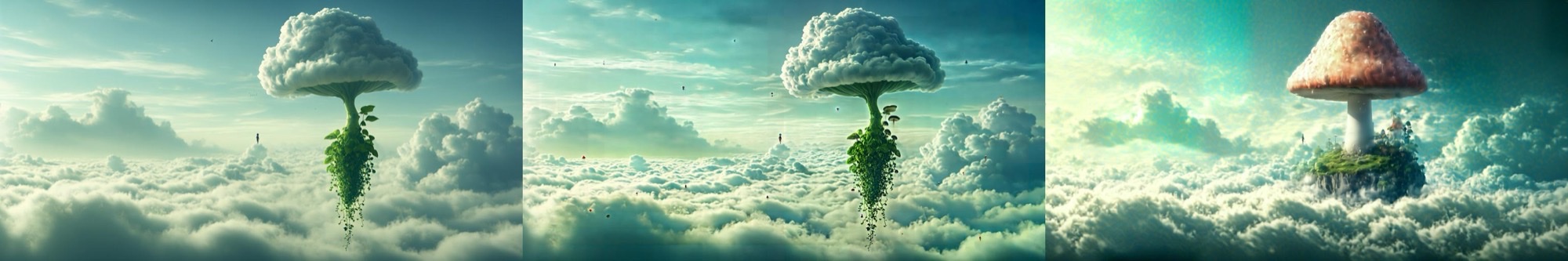}}}  \\
    \\[+0.5em]

    \multicolumn{3}{c}{\makecell{\\\hspace*{0cm} \textbf{SDXL $\times$8} \textit{``lion''} $\rightarrow$ \textit{``tiger''}}}  \\
    \multicolumn{3}{c}{\hspace*{0cm}\raisebox{-0.5\totalheight}{\includegraphics[width=1.0\textwidth]{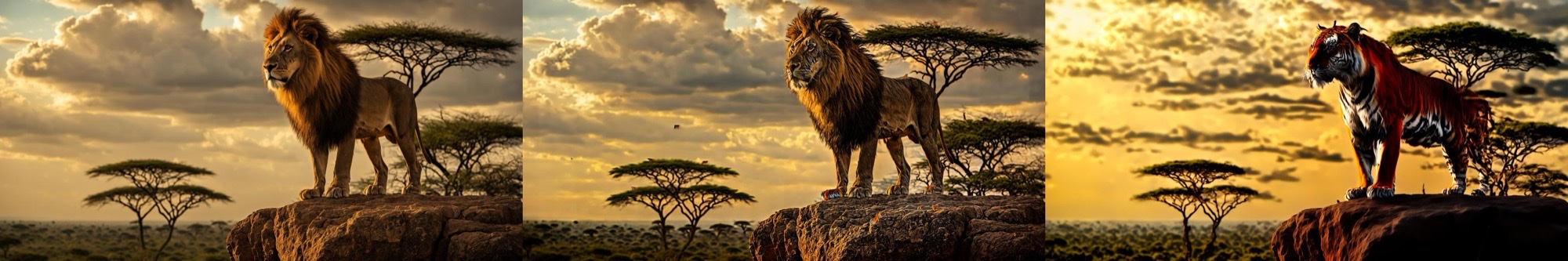}}}  \\
    \\[+0.5em]

    \multicolumn{3}{c}{\makecell{\\\hspace*{0cm} \textbf{SDXL $\times$8} \textit{``shell''} $\rightarrow$ \textit{``crab''}}}  \\
    \multicolumn{3}{c}{\hspace*{0cm}\raisebox{-0.5\totalheight}{\includegraphics[width=1.0\textwidth]{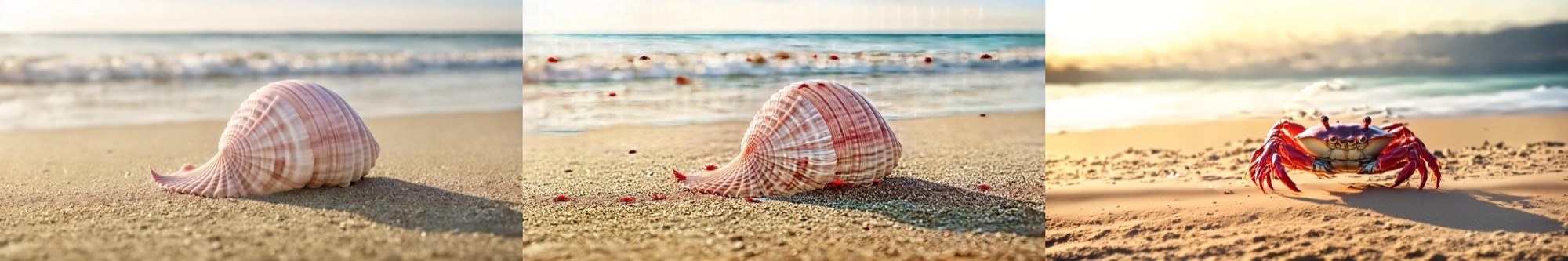}}}  \\
    \\[+0.5em]

    \multicolumn{3}{c}{\makecell{\\\hspace*{0cm} \textbf{SDXL $\times$8} \textit{``snow globe''} $\rightarrow$ \textit{``jungle globe''}}}  \\
    \multicolumn{3}{c}{\hspace*{0cm}\raisebox{-0.5\totalheight}{\includegraphics[width=1.0\textwidth]{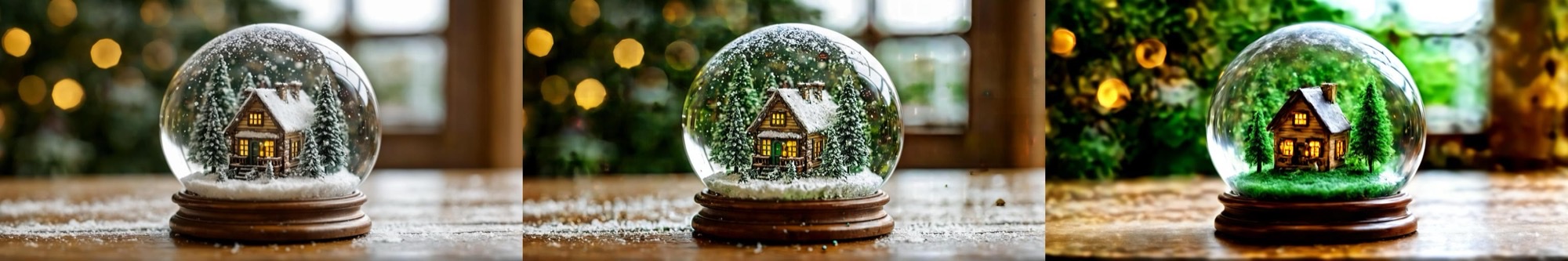}}}  \\
    \\[+0.5em]

    \multicolumn{3}{c}{\makecell{\\\hspace*{0cm} \textbf{SDXL $\times$8} \textit{``whale''} $\rightarrow$ \textit{``turtle''}}}  \\
    \multicolumn{3}{c}{\hspace*{0cm}\raisebox{-0.5\totalheight}{\includegraphics[width=1.0\textwidth]{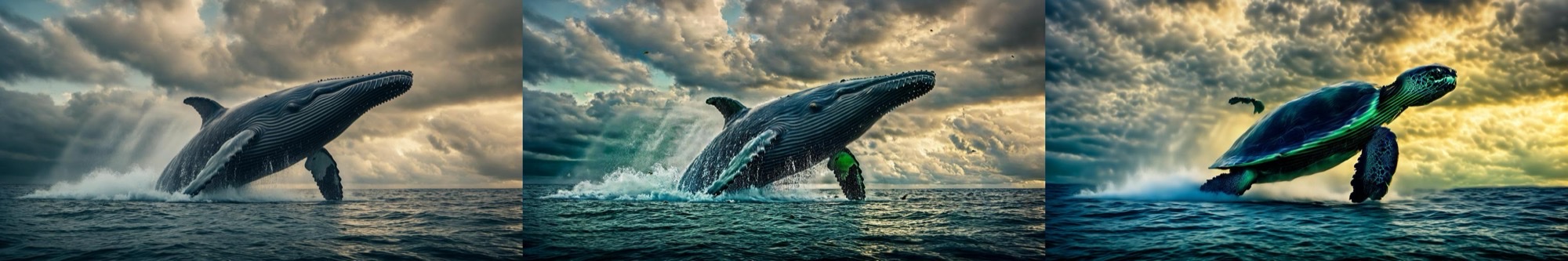}}}  \\
    \\[+0.5em]

    \newpage

    \multicolumn{3}{c}{\makecell{\\\hspace*{0cm} \textbf{SDXL $\times$16} \textit{``apple''} $\rightarrow$ \textit{``pink peach''}}}  \\
    \multicolumn{3}{c}{\hspace*{0cm}\raisebox{-0.5\totalheight}{\includegraphics[width=1.0\textwidth]{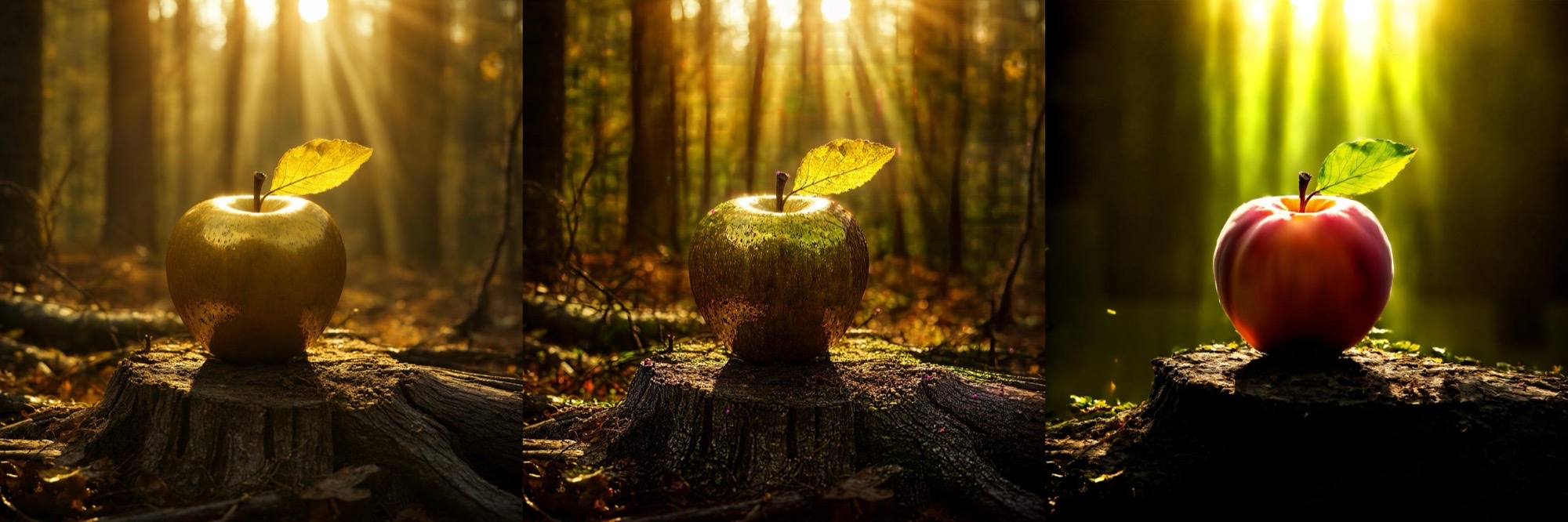}}}  \\
    \\[+0.5em]

    \multicolumn{3}{c}{\makecell{\\\hspace*{0cm} \textbf{SDXL $\times$16} \textit{``bee''} $\rightarrow$ \textit{``hummingbird''}}}  \\
    \multicolumn{3}{c}{\hspace*{0cm}\raisebox{-0.5\totalheight}{\includegraphics[width=1.0\textwidth]{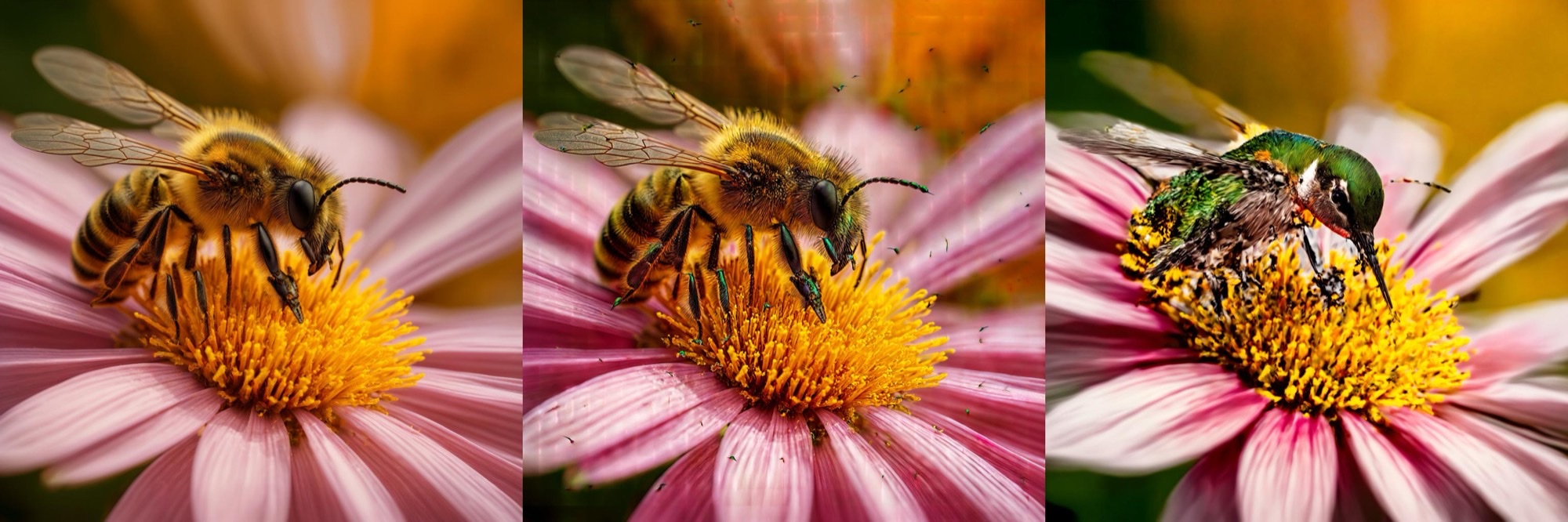}}}  \\
    \\[+0.5em]

    \multicolumn{3}{c}{\makecell{\\\hspace*{0cm} \textbf{SDXL $\times$16} \textit{``bird''} $\rightarrow$ \textit{``owl''}}}  \\
    \multicolumn{3}{c}{\hspace*{0cm}\raisebox{-0.5\totalheight}{\includegraphics[width=1.0\textwidth]{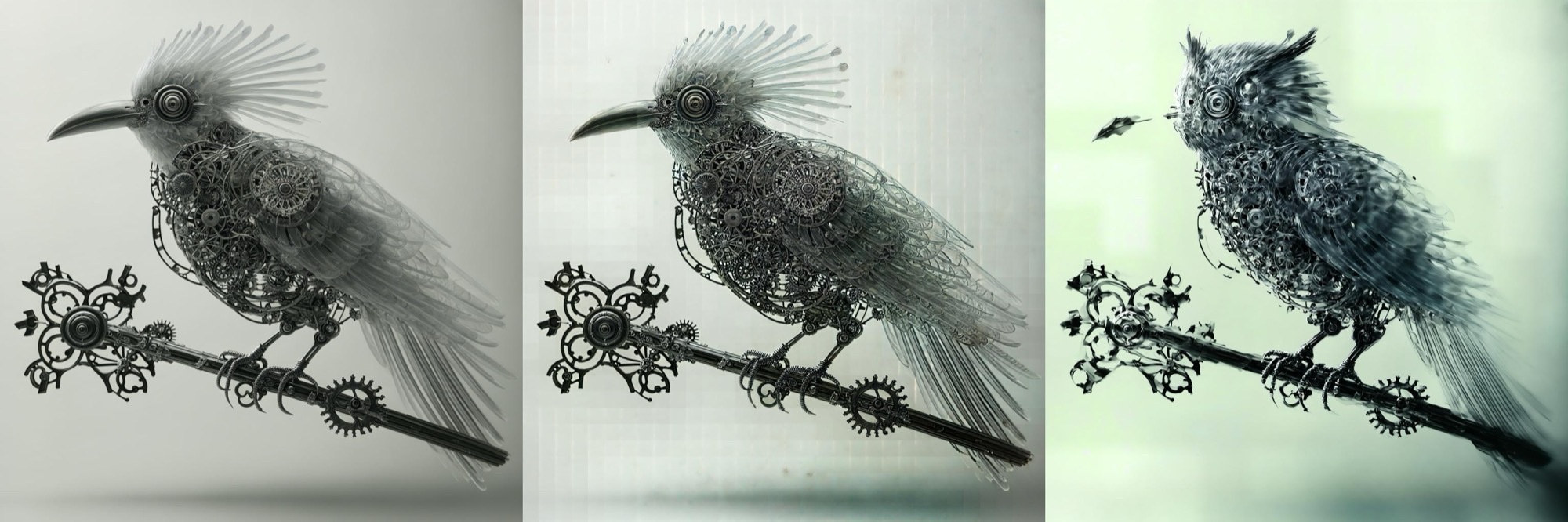}}}  \\
    \\[+0.5em]

    \newpage

    \multicolumn{3}{c}{\makecell{\\\hspace*{0cm} \textbf{SDXL $\times$16} \textit{``mountain''} $\rightarrow$ \textit{``sand dune''}}}  \\
    \multicolumn{3}{c}{\hspace*{0cm}\raisebox{-0.5\totalheight}{\includegraphics[width=1.0\textwidth]{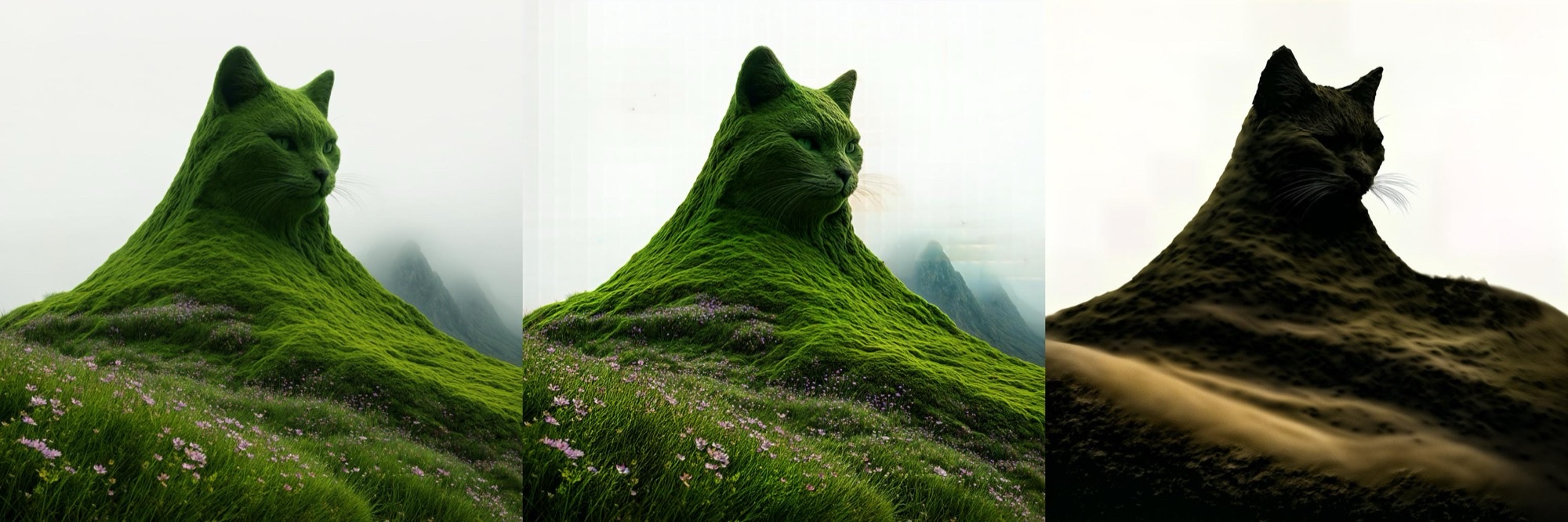}}}  \\
    \\[+0.5em]

    \multicolumn{3}{c}{\makecell{\\\hspace*{0cm} \textbf{SDXL $\times$16} \textit{``stone''} $\rightarrow$ \textit{``Stonehenge''}}}  \\
    \multicolumn{3}{c}{\hspace*{0cm}\raisebox{-0.5\totalheight}{\includegraphics[width=1.0\textwidth]{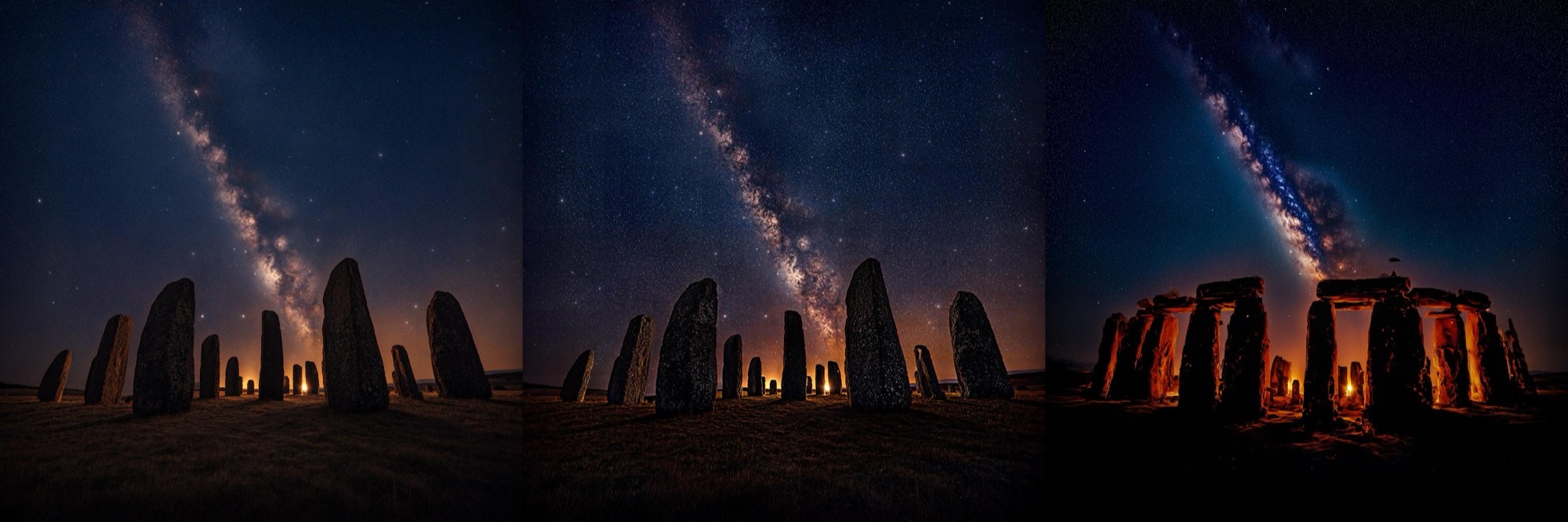}}}  \\
    \\[+0.5em]

    \multicolumn{3}{c}{\makecell{\\\hspace*{0cm} \textbf{SDXL $\times$16} \textit{``waterfall''} $\rightarrow$ \textit{``lava flow''}}}  \\
    \multicolumn{3}{c}{\hspace*{0cm}\raisebox{-0.5\totalheight}{\includegraphics[width=1.0\textwidth]{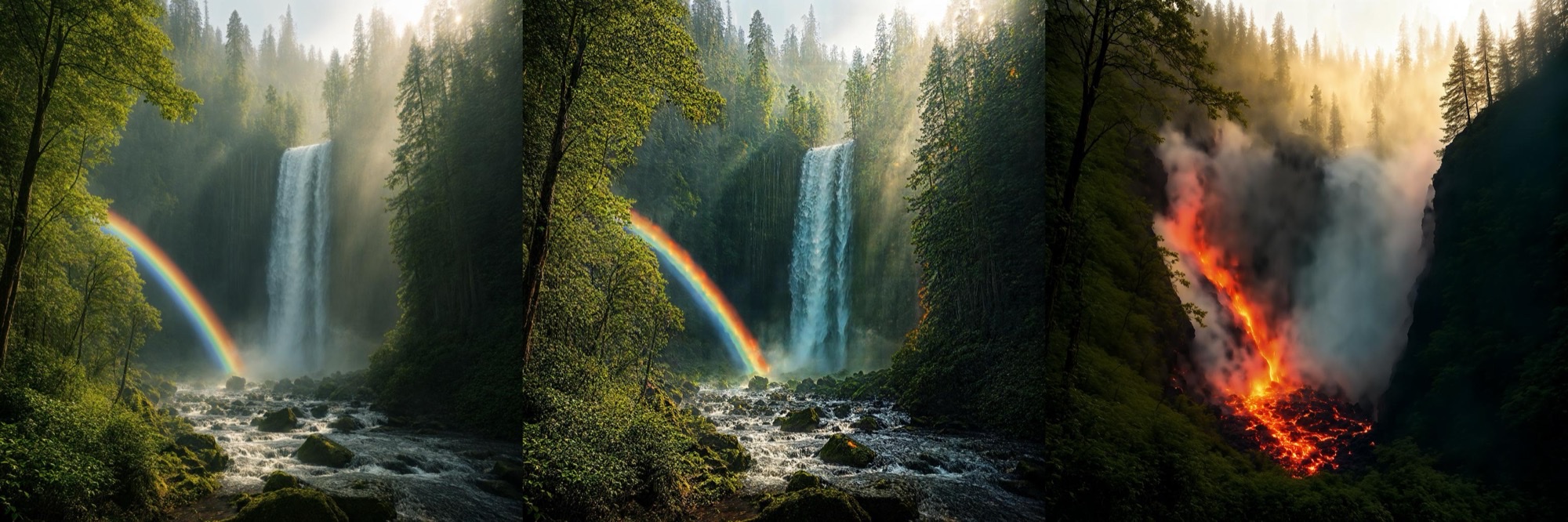}}}  \\
    \\[+0.5em]
    
\end{longtable}
}

\twocolumn

\end{document}